\documentclass{ws-jmrr}

\renewcommand{\draftnote}{} % to remove the timestamp on top-left of every page
\advance\textheight by -\headheight
\advance\textheight by -\headsep

\usepackage{cite}
\usepackage[acronym]{glossaries}
\usepackage{graphicx}
\usepackage{array}
\usepackage{siunitx}         % For consistent unit formatting
\usepackage{makecell}        % For header formatting
\usepackage{comment}
\usepackage{multirow}
\usepackage{tikz}
\usetikzlibrary{backgrounds}
\usetikzlibrary{fit}
\usetikzlibrary{positioning} % for math in positioning?
\usetikzlibrary{calc}
\usetikzlibrary{patterns} % for pattern infill
\usetikzlibrary{quotes} % for text on arrows
\usetikzlibrary{decorations.markings}
\usepackage{pgfplots}
\pgfplotsset{compat=1.16}
\usepgfplotslibrary{fillbetween}

\usetikzlibrary{fit}
\tikzset{
  fitting node/.style={
    inner sep=0pt,
    fill=none,
    draw=none,
    reset transform,
    fit={(\pgf@pathminx,\pgf@pathminy) (\pgf@pathmaxx,\pgf@pathmaxy)}
  },
  reset transform/.code={\pgftransformreset}
}

\usepackage{xspace}
\newcommand*{\eg}{\emph{e.g.}\@\xspace}

\newacronym{rl}{RL}{Reinforcement Learning}
\newacronym{mri}{MRI}{magnetic resonance imaging}
\newacronym{ct}{CT}{Computed Tomography}
\newacronym{us}{US}{ultrasound}
\newacronym{nir}{NIR}{near-infrared}
\newacronym{icg}{ICG}{indocyanine green}
\newacronym{sbr}{SBR}{signal-to-background ratio}
\newacronym{snr}{SNR}{signal-to-noise ratio}
\newacronym{rapn}{RAPN}{robot-assisted partial nephrectomy}
\newacronym{dvrk}{dVRK}{da Vinci Research Kit}
\newacronym{pam}{PAM}{polyacrylamide}
\newacronym{pva}{PVA}{polyvinyl alcohol}
\newacronym{pla}{PLA}{polylactic acid}
\newacronym{dsc}{DSC}{DICE similarity coefficient}
\newacronym{mis}{MIS}{minimally-invasive surgery}

\usepackage{siunitx}
\begin{document}

%\catchline{0}{0}{2013}{}{}

\markboth{Kilmer et al.}{Towards Fluorescence-Guided Autonomous Robotic Partial Nephrectomy on Novel Realistic Hydrogel-Based Phantoms}

\title{Towards Fluorescence-Guided Autonomous Robotic Partial Nephrectomy on Novel Tissue-Mimicking Hydrogel Phantoms}

\author{
Ethan Kilmer$^{a \dagger}$,
Joseph Chen$^{a \dagger}$,
Jiawei Ge$^a$,
Preksha Sarda$^b$,
Richard Cha$^c$,
Kevin Cleary$^c$,\\
Lauren Shepard$^d$,
Ahmed Ezzat Ghazi$^d$,
Paul Maria Scheikl$^a$,
Axel Krieger$^a$
}

\address{$^a$ Laboratory of Computational Sensing and Robotics, Johns Hopkins University, Baltimore, MD 21218, USA \\ E-mail: pscheik1@jhu.edu}
\address{$^b$ Department of Computer Science and Software Engineering, Rose-Hulman Institute of Technology, Terre Haute, IN 47803, USA}
\address{$^c$ Sheikh Zayed Institute for Pediatric Surgical Innovation, Children’s National Hospital, Washington, DC 20010, USA}
\address{$^d$ Department of Urology, Johns Hopkins University, Baltimore, MD 21218, USA}
\address{$^\dagger$ Equal contribution.}

\maketitle

\thispagestyle{headings} % move the page 1 sign from bottom to top-right.

\begingroup
  \renewcommand{\thefootnote}{} % suppress marker
  \footnotetext{Preprint of an article accepted for publication in the Journal of Medical Robotics Research, 2025. Copyright World Scientific Publishing Company [https://worldscientific.com/worldscinet/jmrr].}
\endgroup

\begin{abstract}
Autonomous robotic systems hold potential for improving renal tumor resection accuracy and patient outcomes.
We present a fluorescence-guided robotic system capable of planning and executing incision paths around exophytic renal tumors with a clinically relevant resection margin.
Leveraging point cloud observations, the system handles irregular tumor shapes and distinguishes healthy from tumorous tissue based on near-infrared imaging, akin to indocyanine green staining in partial nephrectomy.
Tissue-mimicking phantoms are crucial for the development of autonomous robotic surgical systems for interventions where acquiring ex-vivo animal tissue is infeasible, such as cancer of the kidney and renal pelvis.
To this end, we propose novel hydrogel-based kidney phantoms with exophytic tumors that mimic the physical and visual behavior of tissue, and are compatible with electrosurgical instruments, a common limitation of silicone-based phantoms.
In contrast to previous hydrogel phantoms, we mix the material with near-infrared dye to enable fluorescence-guided tumor segmentation.
Autonomous real-world robotic experiments validate our system and phantoms, achieving an average margin accuracy of \SI{1.44}{\mm} in a completion time of \SI{69}{\sec}.
\end{abstract}


\begin{multicols}{2}
\section{Introduction}
More than 65,000 patients are diagnosed with cancers of the kidney and renal pelvis every year, leading to approximately 15,000 deaths in the US alone~\cite{Siegel2019-iu}. 
Partial nephrectomy is the accepted standard of care for localized small renal masses.
In contrast to traditional radical nephrectomy, where the whole kidney is removed, partial nephrectomy only removes the affected tumorous tissue plus a margin of healthy tissue.
When choosing the resection margin, surgeons must strike a balance, ensuring that all cancerous cells are removed to minimize tumor recurrence while sparing as much healthy tissue as possible to maximize organ function and recovery.

Autonomous robotic tumor resection, where the robot plans and executes tumor resection under human supervisory approval, offers a promising solution to improve tumor margin accuracy.
In the development of autonomous surgical robot systems for tumor resection, realistic surgical ph-
\def\imgheight{0.15\columnwidth}
\def\imgsep{0.2mm}
\begin{figurehere}
    \begin{center}
        \begin{tikzpicture}
        \node (system) {\includegraphics[width=0.8\columnwidth]{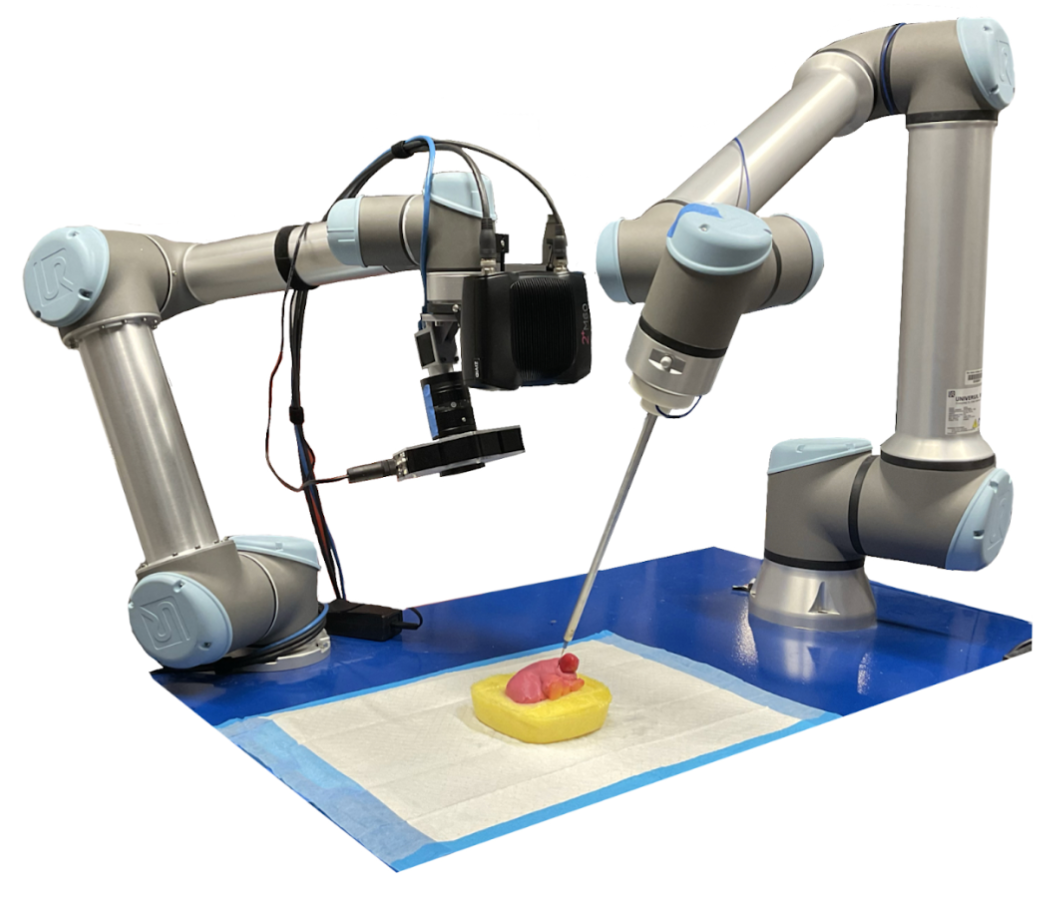}};
        
        \node[align=center] (label zivid) at (0.5, 2.5) {RGBD\\ Camera};
        \node[align=center] (label nir) at (-2.5, 2.2) {NIR\\ Camera};
        \node[align=center] (label light) at (-2.5, -2.5) {NIR\\ Light};
        \node[align=center] (label kidney) at (0.5, -3.0) {Kidney\\ Phantom};
        \node[align=center] (label cautery) at (2.5, -2.5) {Electrocautery\\ Instrument};

        \draw[->, ultra thick] (label zivid.south) -- (-0.0, 1.5);
        \draw[->, ultra thick] (label nir.south) -- (-0.9, 0.8);
        \draw[->, ultra thick] (label light.east) -- (-0.9, -0.3);
        \draw[->, ultra thick] (label kidney.north) -- (0.2, -1.7);
        \draw[->, ultra thick] (label cautery.north west) -- (0.4, -1.2);
        
        \end{tikzpicture}
        \caption{Robot system for autonomous kidney tumor resection.}
        \label{fig1}
    \end{center}
\end{figurehere}
\def\imgsep{0.2mm}
\def\colimgwidth{0.15\textwidth}
\begin{figure*}
    \begin{center}
        \begin{tikzpicture}
        \node (phantom 1) {\includegraphics[height=\colimgwidth]{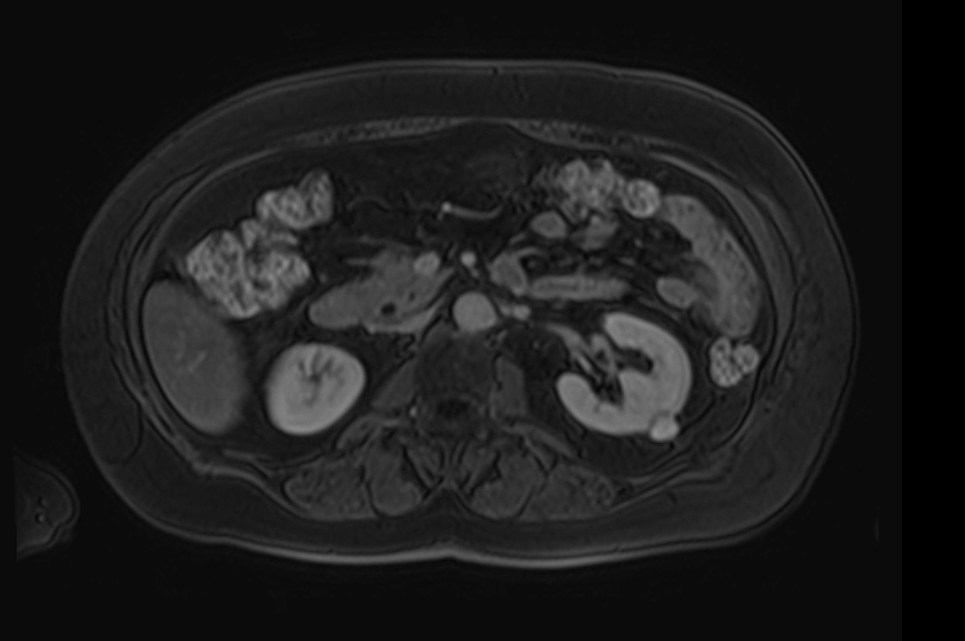}};
        \node[right = 4mm of phantom 1.east, anchor = west] (phantom 2) {\includegraphics[height=\colimgwidth]{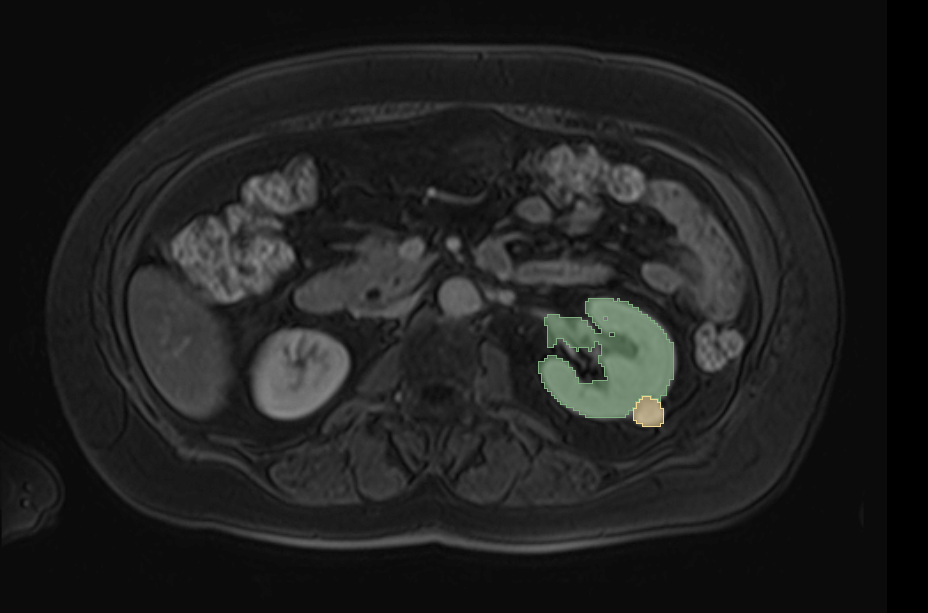}};
        \node[right = 4mm of phantom 2.east, anchor = west] (phantom 3) {\includegraphics[height=\colimgwidth]{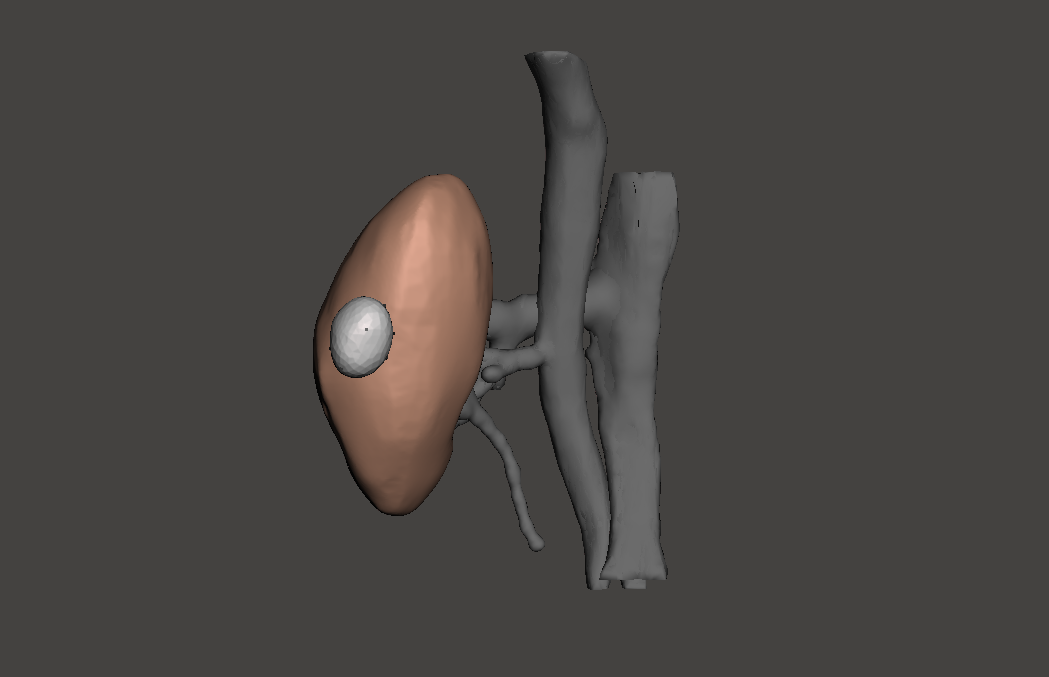}};
        \node[right = 4mm of phantom 3.east, anchor = west] (phantom 4) {\includegraphics[height=\colimgwidth]{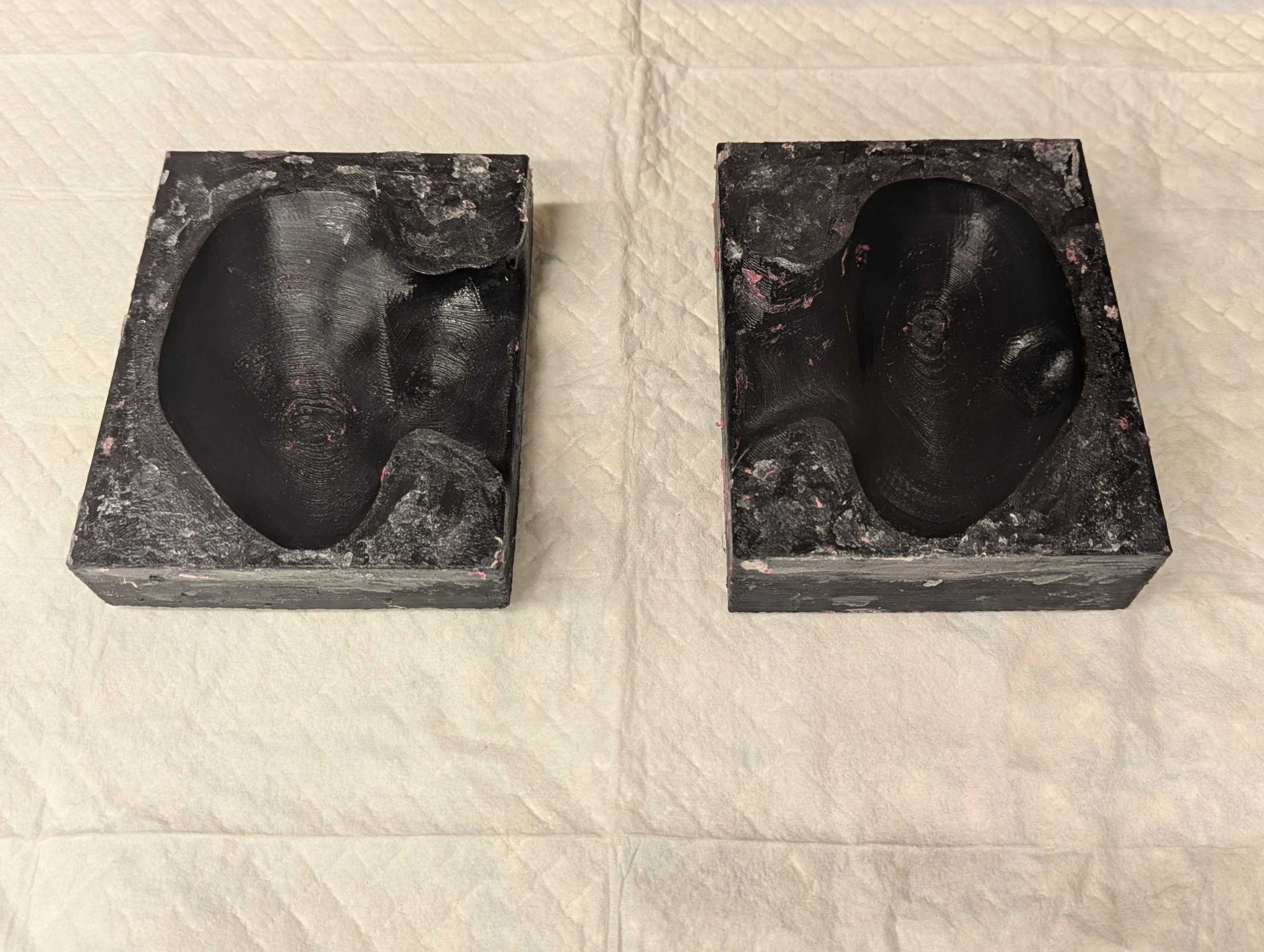}};

        \node[below = 0.1cm of phantom 1.south, anchor=north, align=center] {Patient CT};
        \node[below = 0.1cm of phantom 2.south, anchor=north, align=center] {Segmentation};
        \node[below = 0.1cm of phantom 3.south, anchor=north, align=center] {Volumetric Meshes};
        \node[below = 0.1cm of phantom 4.south, anchor=north, align=center] {Mold};

        \draw[->, ultra thick] (phantom 1.east) -- (phantom 2.west);
        \draw[->, ultra thick] (phantom 2.east) -- (phantom 3.west);
        \draw[->, ultra thick] (phantom 3.east) -- (phantom 4.west);
         
        \end{tikzpicture}
        \caption{
        Patient CT data is segmented to create volumetric meshes of the kidney, embedded tumor, and the renal hilum.
        After smoothing the meshes, the negatives of the volumetric meshes are 3D printed to serve as molds for our hydrogel phantoms.}
        \label{fig:manufacturing}
    \end{center}
\end{figure*}
antoms play a crucial role, since obtaining ex-vivo animal tissue phantoms with specific kidney tumors is not feasible.
Realistic phantoms contain tumors to match patient specific data and replicate relevant aspects of real surgical interventions, including the ability to resect phantom tissue through electrocautery.
Accurate tumor localization from image observations is the foundational perception skill that is required for subsequent planning and execution of robotic tumor resection trajectories.
Surgeons often inject a \gls{nir} fluorescent marker such as \gls{icg} to improve the visual contrast between tumor and kidney.

To advance autonomous robotic tumor resection, McKinley et al. use the \gls{dvrk} to palpate and remove cylindrical rubber shapes embedded in a silicone rubber pad~\cite{mckinley2016interchangeable}.
Hu et al. perform semi-autonomous tumor debridement of jelly-like mixtures with a suction tool~\cite{hu2018semi}.
While these advances show a proof of concept, they are challenging to extend to realistic surgical settings due to their reliance on simplified phantom models that are not patient specific.
Recently, Ge et al. achieved autonomous resection of a non-planar pseudotumor chip using porcine tongue samples~\cite{ge2024autonomous}, and Marahrens et al. demonstrated ultrasound-guided surgical margin marking for spherical, hydrated \gls{pam} pseudotumor beads in a porcine liver~\cite{marahrens2024ultrasound}.
However, neither study uses a realistic 3D irregularly shaped tumor as the target.

Recent progress in medical image segmentation and 3D printing techniques has allowed researchers to print deformable material organ replicas based on patient-specific data~\cite{wang2017review, nieva2024developing}. 
Nieva-Esteve et al. present a novel family of silicone gel-based inks that can be used to 3D print organ replicas with greater mechanical similarity to real tissue than traditional materials~\cite{nieva2024developing}. 
However, an important requirement for tumor resection phantoms is mimicking dissection behavior under electrocautery, a critical limitation to common silicone phantoms.
Amiri et al. present tissue-mimicking phantom materials based on fat, water, and agar/gelatin that are electrically-conductive, thus supporting electrosurgery~\cite{Amiri2022-rg}.
In a related effort, Melynk et al. presented a novel hydrogel kidney phantom for partial nephrectomy.
However, this phantom does not support fluorescence that mimics \gls{icg}~\cite{melnyk2020mechanical}.\\

In this work, we propose a fluorescence-guided autonomous robotic partial nephrectomy system, illustrated in Fig.~\ref{fig1}, and demonstrate its tumor resection capabilities on tissue-mimicking kidney phantoms that feature 3D irregularly shaped tumors.
Our contributions are 1) the adaption of novel hydrogel-based patient-specific phantoms for partial nephrectomy to \gls{nir} visibility, and 2) a system for autonomously segmenting tumor and kidney to plan and execute a circumferential surface incision around the embedded tumor, respecting a realistic surgical margin.
The segments of the phantoms are visible under \gls{ct} and contain \gls{nir} fluorescent dye for image-based segmentation.
Specifically, our phantoms enable reproducing \gls{nir} imagining capability with adjustable \gls{sbr} that enables better visualization of tumor edges and margins~\cite{abaza2017differential, henrich2024tracking, melnyk2020mechanical}.
We perform autonomous robotic incisions on four (N=4) hydrogel phantoms, validating the use of these phantoms and representing a first crucial step in the tumor resection procedure.
\begin{figurehere}
    \begin{center}
        \begin{tikzpicture}
            \node[inner sep = 0, outer sep = 0] (first) {\includegraphics[width=0.33\columnwidth]{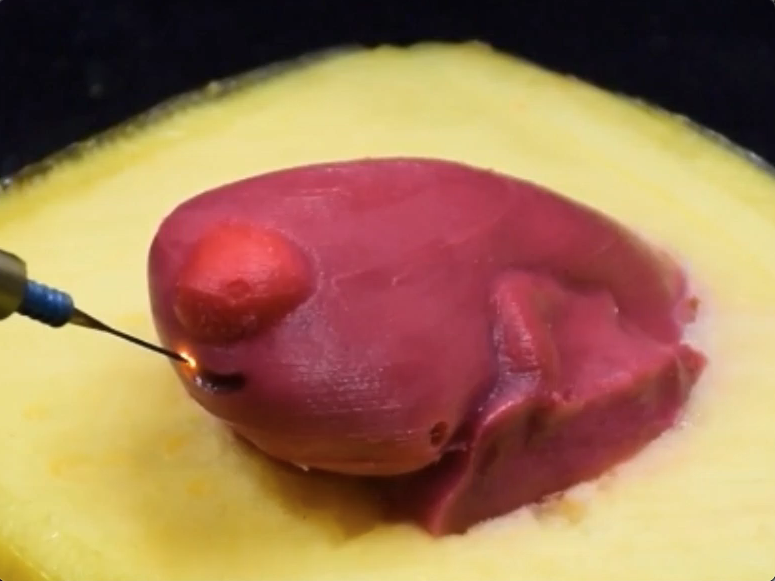}};
            \node[inner sep = 0, outer sep = 0, right = 1mm of first.east, anchor=west] (second) {\includegraphics[width=0.33\columnwidth]{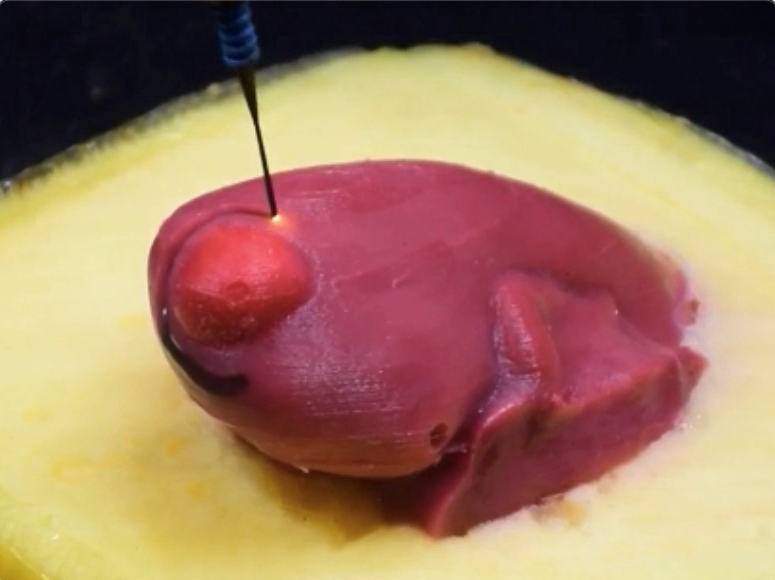}};
            \node[inner sep = 0, outer sep = 0, right = 1mm of second.east, anchor=west] (third) {\includegraphics[width=0.33\columnwidth]{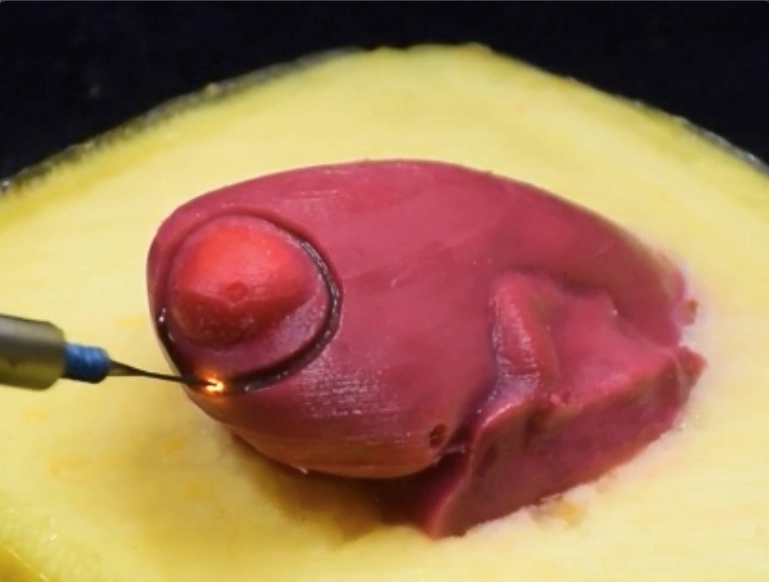}};
        \end{tikzpicture}
        \caption{Hydrogel phantoms are compatible with electrosurgical instruments, such as cauterization electrodes, providing more realistic simulation capabilities compared to silicone phantoms. This enables autonomous robotic electrosurgical incisions.}
        \label{fig:samplecut}
    \end{center}
\end{figurehere}

\section{Methods}
Progress in autonomous robotic tumor resection relies on three components: realistic phantoms, accurate tumor localization, and reliable planning and execution of resection paths.
To this end, we design \gls{nir} hydrogel kidney phantoms based on segmented patient data for tissue-mimicking partial nephrectomy samples.
We propose a perception pipeline to autonomously capture and segment the surgical scene based on point cloud observations.
The segmented point cloud is subsequently used to plan a robotic incision path around the tumor on the kidney phantom.
The incision is executed on the real robotic system with an electrocautery instrument as shown in Fig.~\ref{fig:samplecut}.

\subsection{Kidney Phantom}
We adapt previously developed tissue-mimicking kidney phantoms~\cite{melnyk2020mechanical, Ghazi2021-ye} to include fluorescence for \gls{nir} imaging support.
Instead of positive staining \gls{icg}, we seek to replicate the negative staining technique where fluorescent dye is injected into the bloodstream causing healthy tissue to illuminate under \gls{nir} light, while tumorous tissue remains dark.
This method has been clinically validated, with one study reporting differential fluorescence in $65$ of $79$ ($82\%$) of renal tumors~\cite{angell2013optimization}.
The Da Vinci Firefly imaging system (Intuitive Surgical, Sunnyvale, CA) is commonly employed in surgical procedures to visualize the negative staining effect~\cite{Krane2012-wn}.

Our tissue-mimicking kidney phantoms are created from anonymized patient data.
We follow previous work to first segment kidney and tumor from patient DICOM data~\cite{melnyk2020mechanical}.
The segmented data is then converted into mesh files to create molds for the kidney phantoms with a boolean difference operation.
The molds are printed in \gls{pla} on a Ender-3 Max Neo (Creality,  Shenzhen, China). 

The kidney phantoms are fabricated using a hydrogel material that is created by first mixing water with \gls{pva} and then repeatedly freezing and thawing the mixture.
By adjusting the concentration of \gls{pva} and the number of freeze thaw cycles, hydrogel is capable of mimicking tumor and kidney tissues~\cite{melnyk2020mechanical}, and even allows for cutting with electrosurgical instruments.
Two batches of 10\% \gls{pva} are used to model the tumor and healthy kidney tissue.
2 w/v\% barium sulfate (BaSO$_4$, Sigma-Aldrich, St. Louis, MO) is added to the tumor batch for contrast enhancement under \gls{ct}.
The kidney batch is mixed with 0.44 w/v\% \gls{nir} fluorescence (3-Hour IR ChemLight, Cylume Technologies, Springfield, MA) to reproduce the visual behavior of intraoperative negative staining \gls{icg} injection.
1.25 w/v\% acrylic white paint is added as a bonding agent, as the \gls{nir} dye does not mix well with water.
The mixture is injected into the mold and undergoes overnight freeze-thaw cycles for solidification.

We create a total number of four kidney phantom models from anonymized patient data that show a variety of kidney and tumor size as well as anatomical sites.
The fabricated phantoms are shown in the first column of Fig.~\ref{fig:models}.
The anatomical site and size of the tumors are 
1) right medial, \SI{17}{\mm},
2) right anterior upper pole, \SI{35}{\mm},
3) right upper pole, \SI{20}{\mm}, and
4) right medial, \SI{30}{\mm}.

\def\imgsep{0.3cm}
\def\colimgwidth{0.12\textwidth}
\begin{figure*}
    \begin{center}
        \begin{tikzpicture}
        \node (phantom 1) {\includegraphics[height=\colimgwidth]{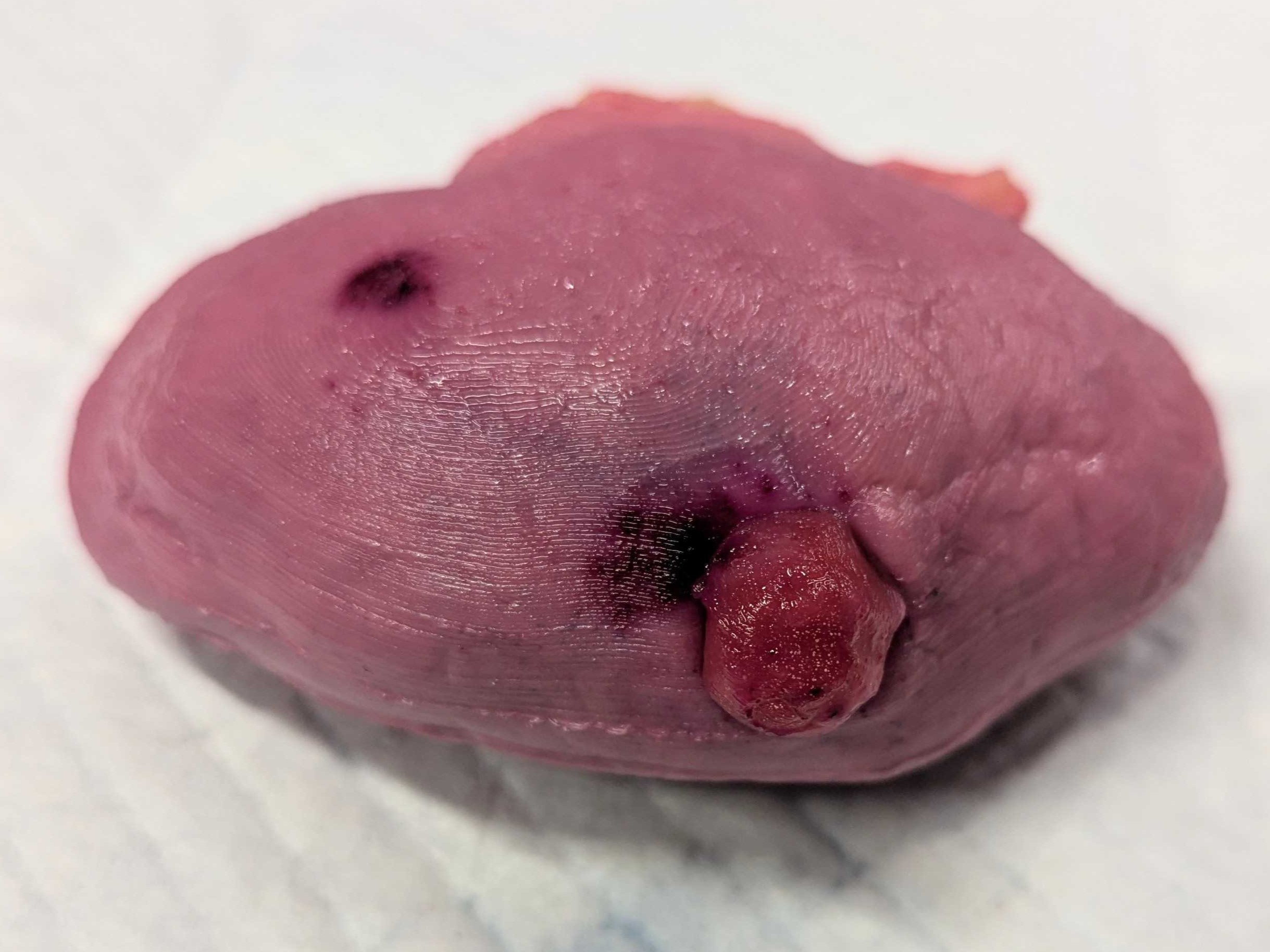}};
        \node[below = 0cm of phantom 1.south, anchor = north] (phantom 2) {\includegraphics[height=\colimgwidth]{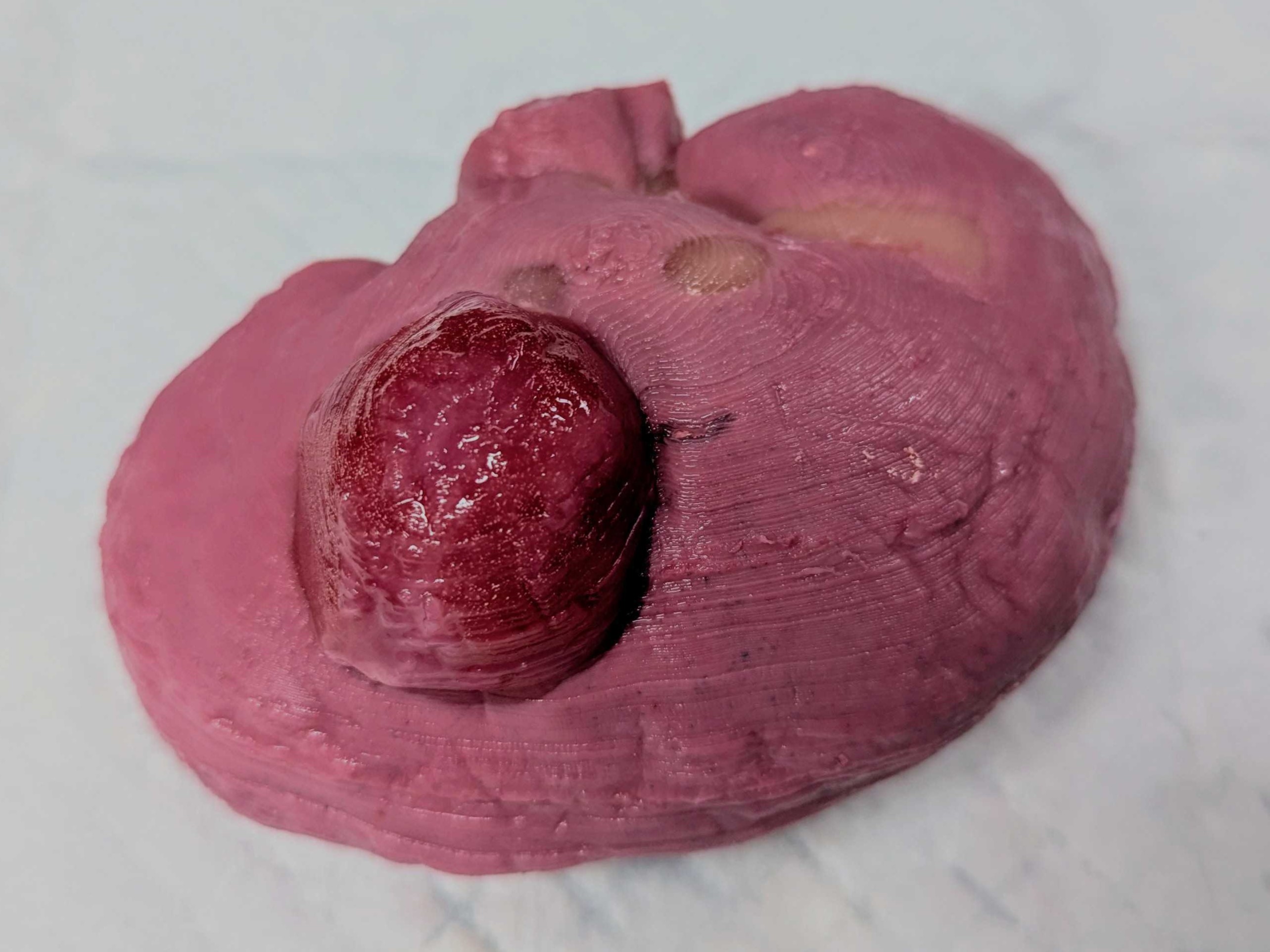}};
        \node[below = 0cm of phantom 2.south, anchor = north] (phantom 4) {\includegraphics[height=\colimgwidth]{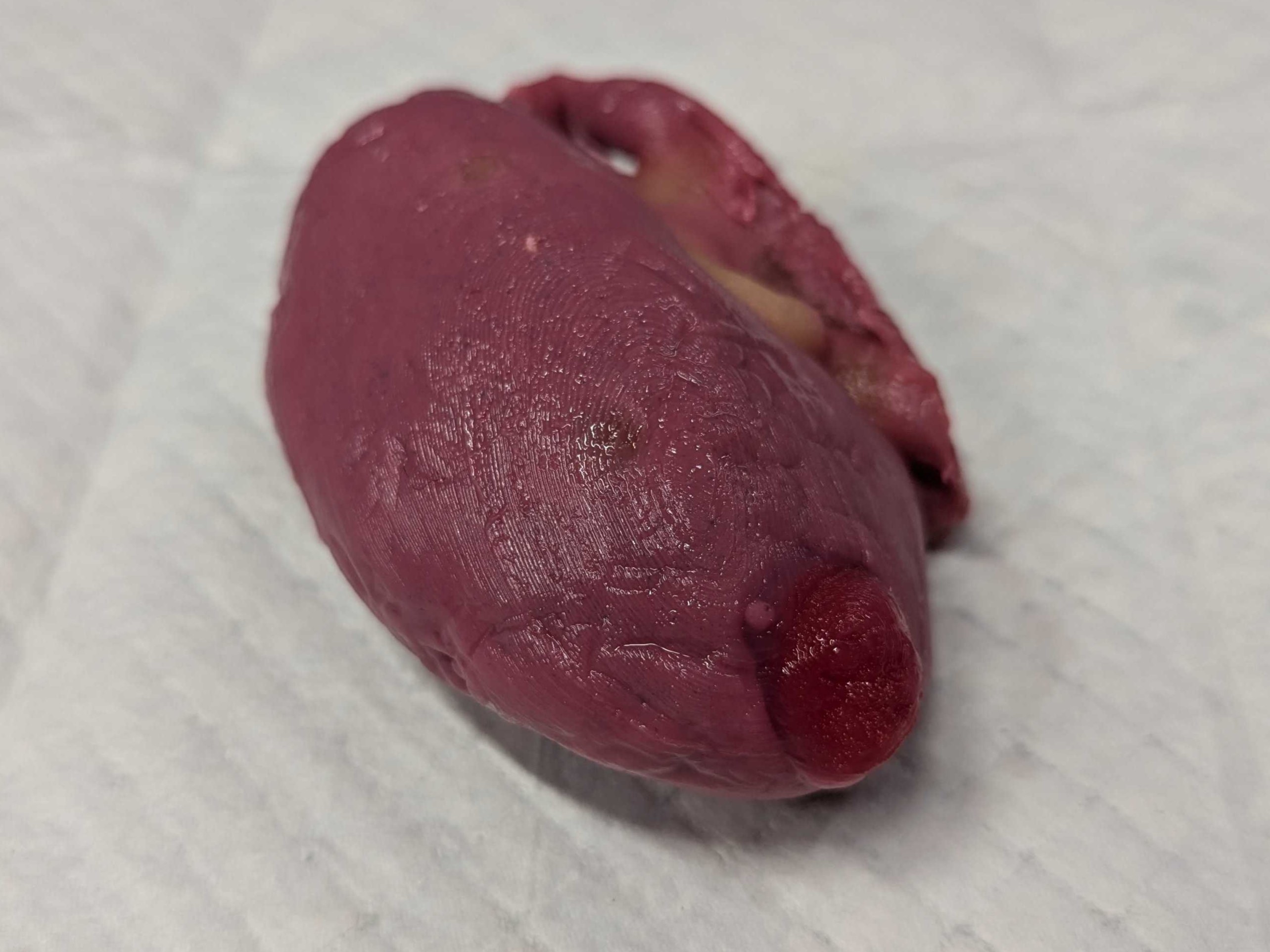}};
        \node[below = 0cm of phantom 4.south, anchor = north] (phantom 5) {\includegraphics[height=\colimgwidth]{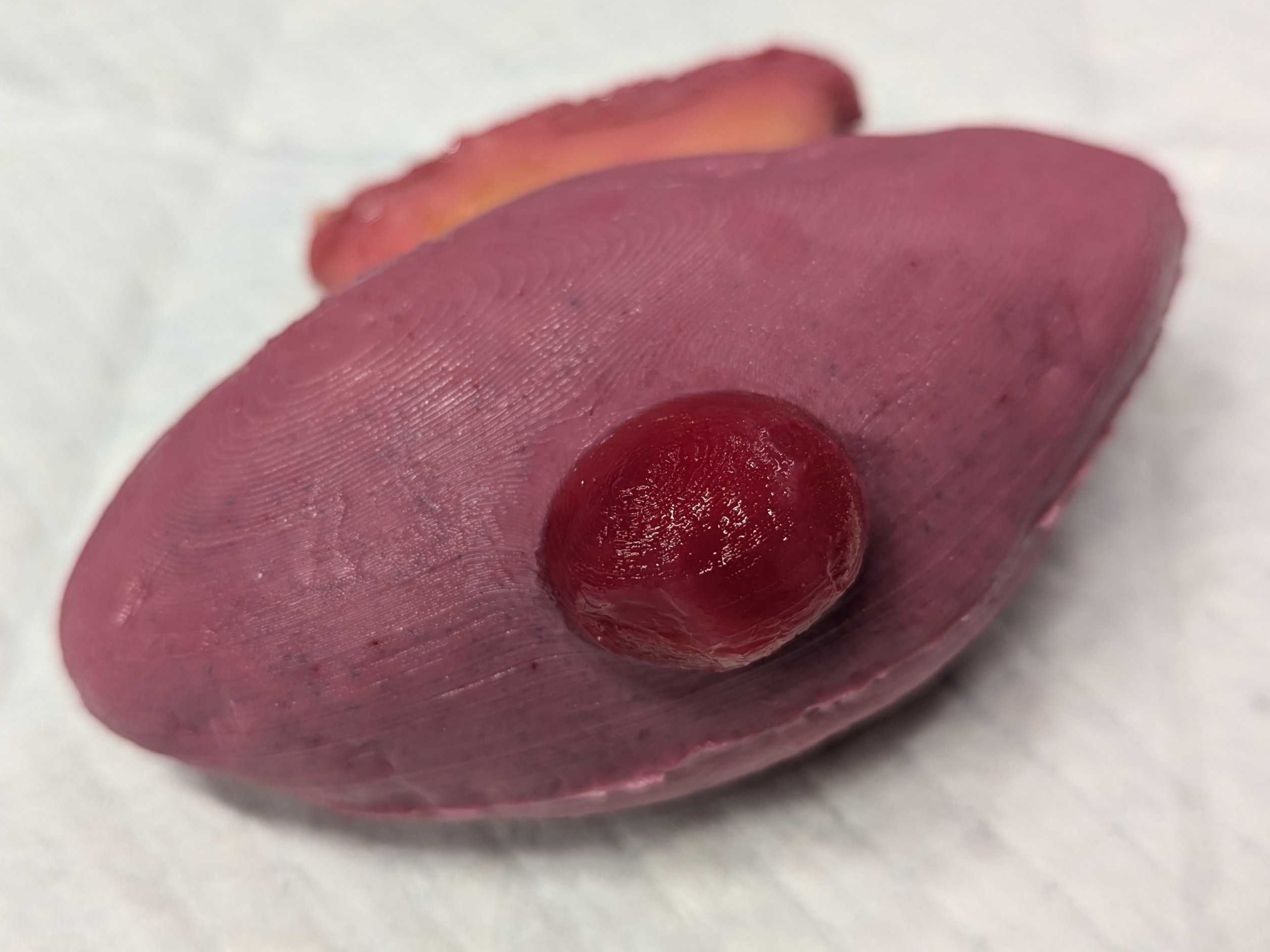}};
        
        \node[right = \imgsep of phantom 1.east, anchor=west] (phantom 1 nir) {\includegraphics[height=\colimgwidth]{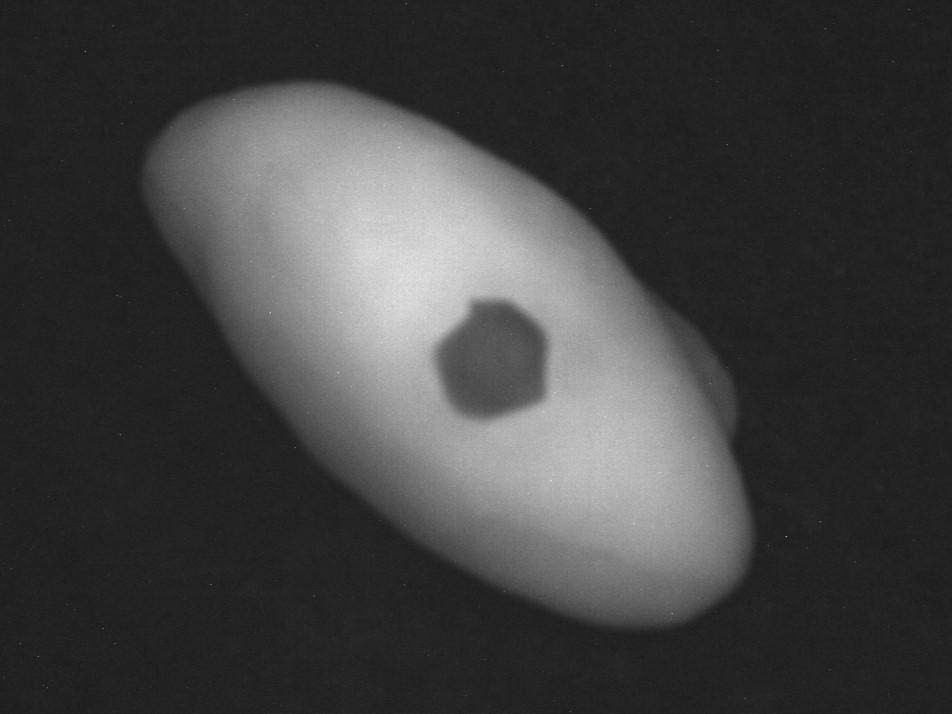}};
        \node[below = 0cm of phantom 1 nir.south, anchor = north] (phantom 2 nir) {\includegraphics[height=\colimgwidth]{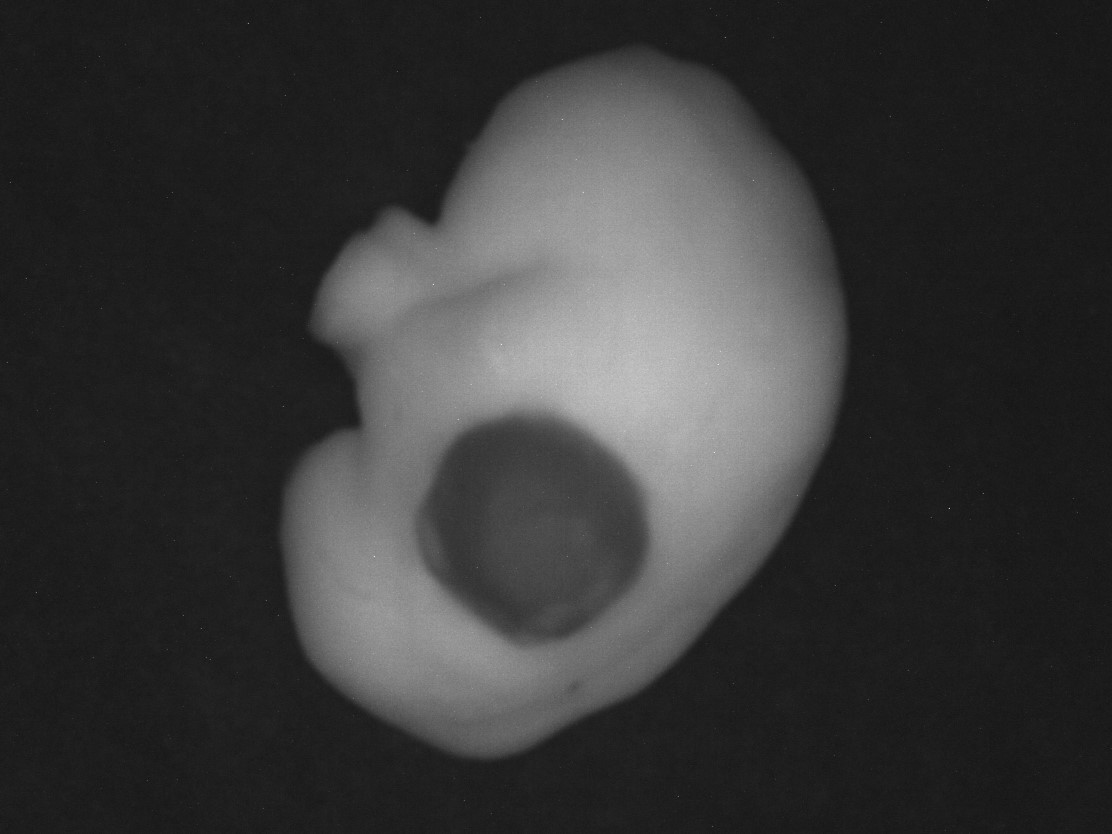}};
        \node[below = 0cm of phantom 2 nir.south, anchor = north] (phantom 4 nir) {\includegraphics[height=\colimgwidth]{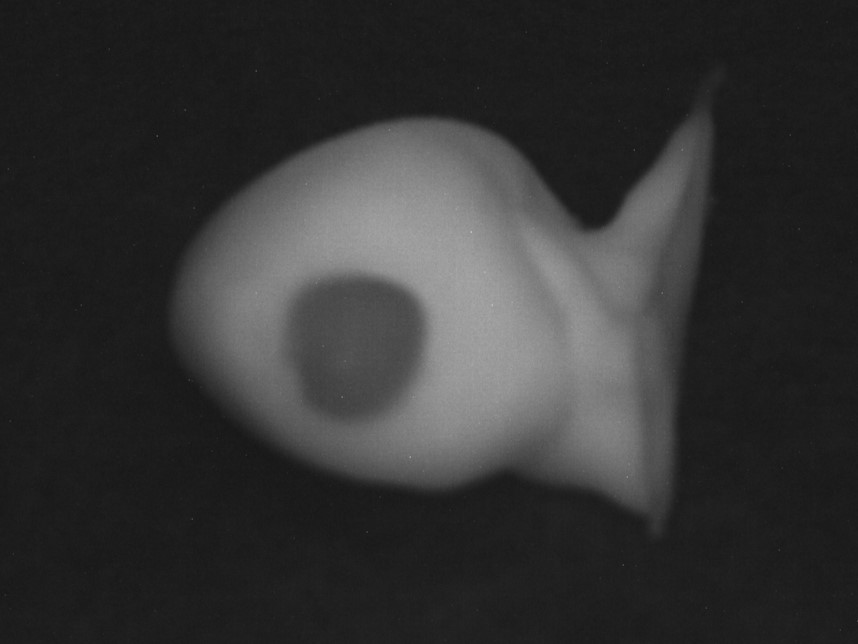}};
        \node[below = 0cm of phantom 4 nir.south, anchor = north] (phantom 5 nir) {\includegraphics[height=\colimgwidth]{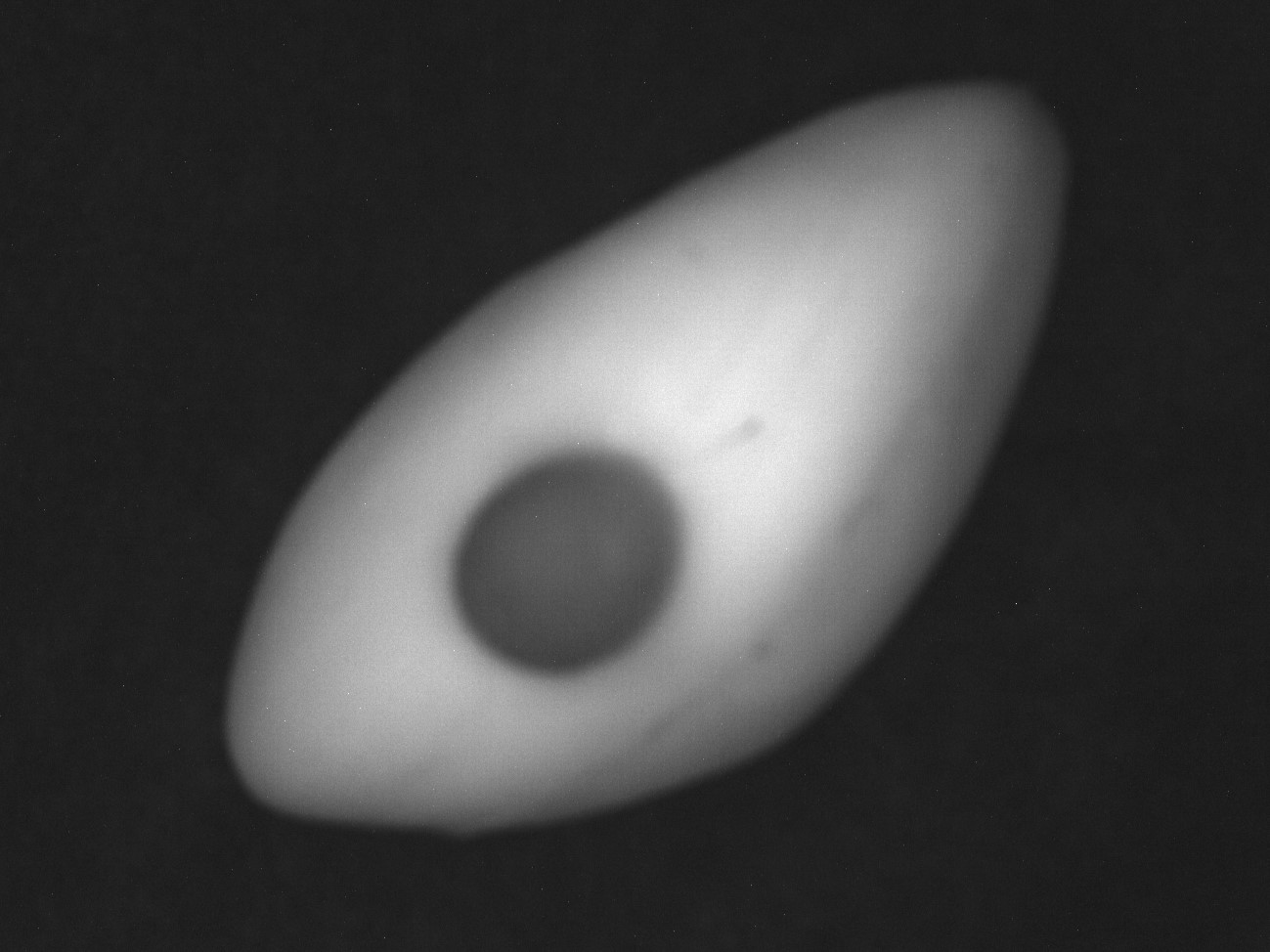}};
        
        \node[right = \imgsep of phantom 1 nir.east, anchor=west] (phantom 1 sam) {\includegraphics[height=\colimgwidth]{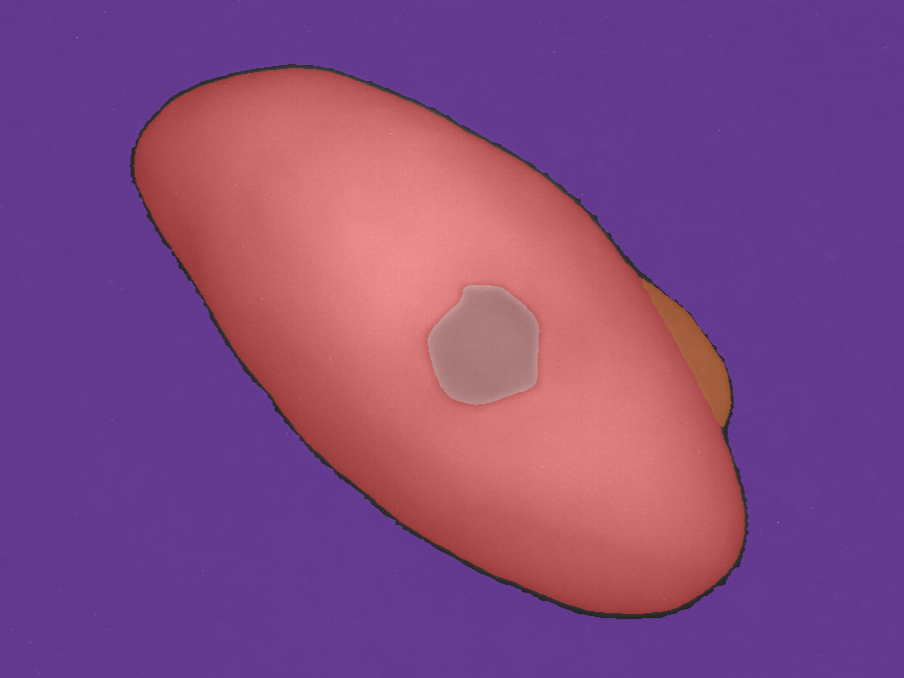}};
        \node[below = 0cm of phantom 1 sam.south, anchor = north] (phantom 2 sam) {\includegraphics[height=\colimgwidth]{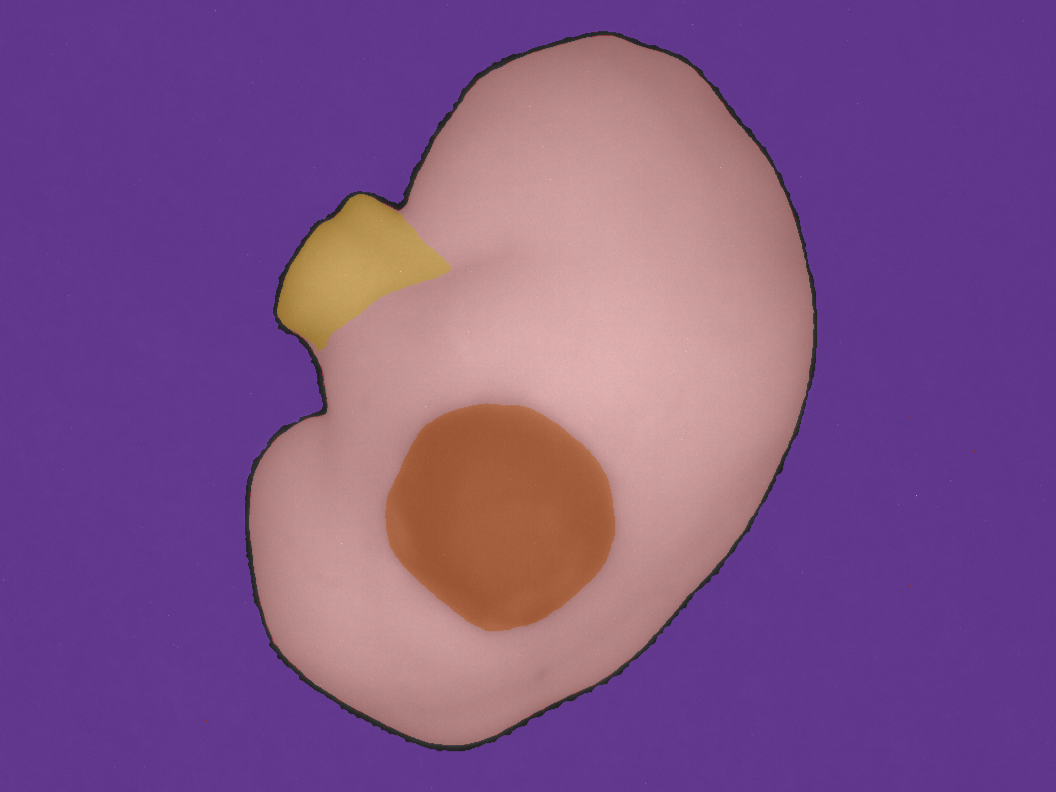}};
        \node[below = 0cm of phantom 2 sam.south, anchor = north] (phantom 4 sam) {\includegraphics[height=\colimgwidth]{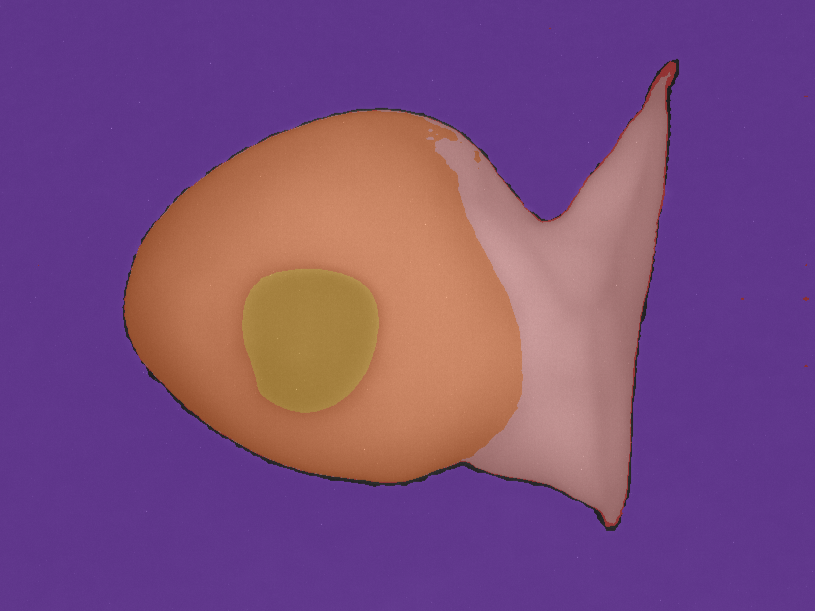}};
        \node[below = 0cm of phantom 4 sam.south, anchor = north] (phantom 5 sam) {\includegraphics[height=\colimgwidth]{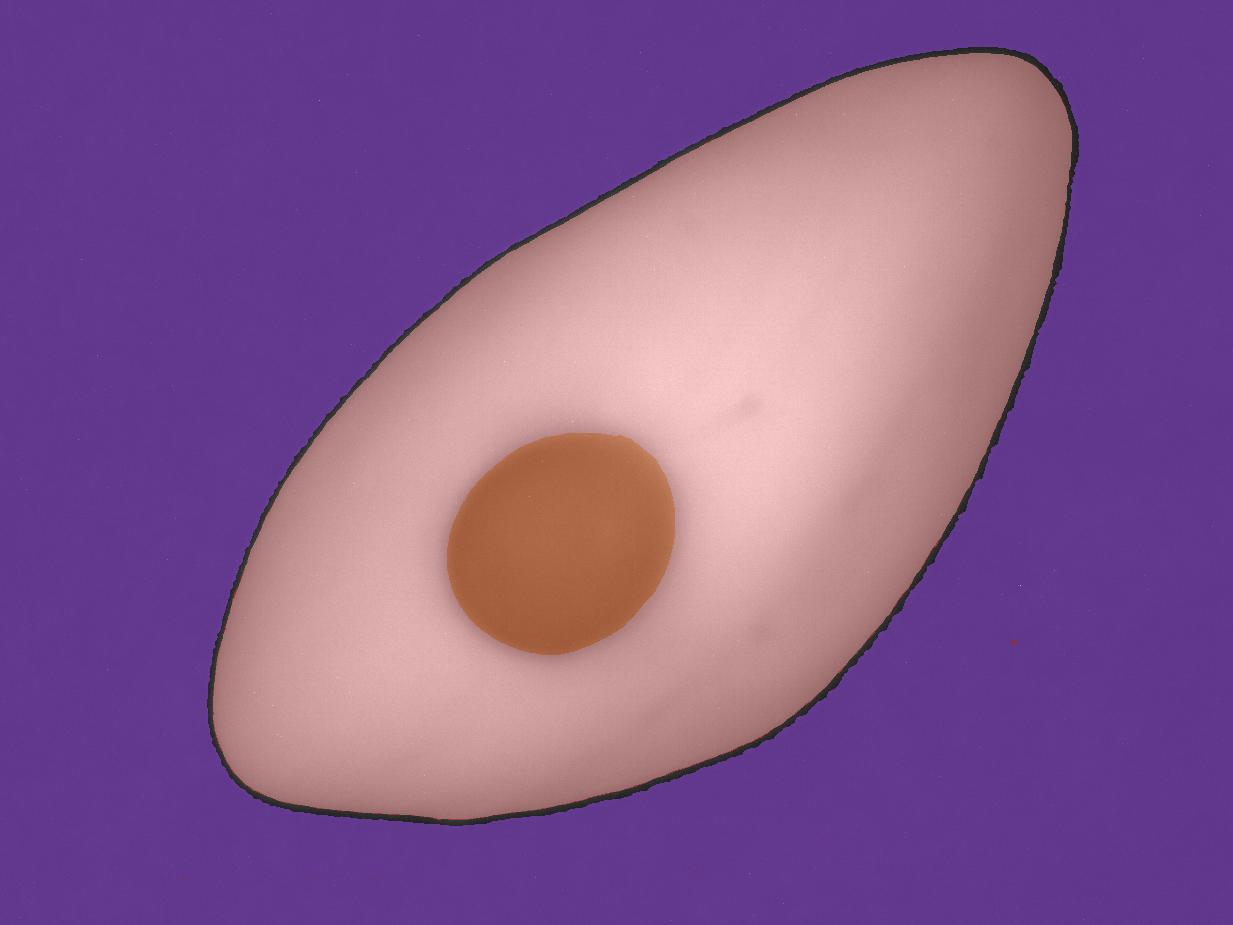}};
        
        \node[right = \imgsep of phantom 1 sam.east, anchor=west] (phantom 1 path) {\includegraphics[height=\colimgwidth]{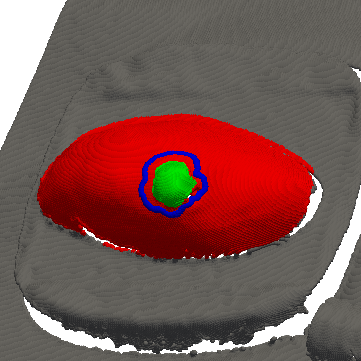}};
        \node[below = 0cm of phantom 1 path.south, anchor = north] (phantom 2 path) {\includegraphics[height=\colimgwidth]{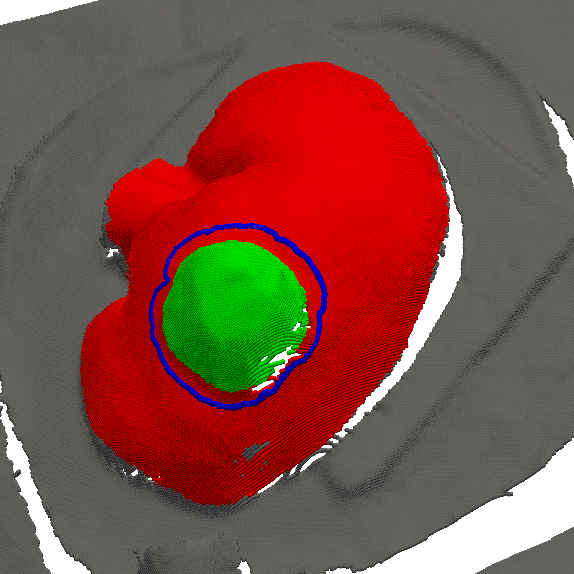}};
        \node[below = 0cm of phantom 2 path.south, anchor = north] (phantom 4 path) {\includegraphics[height=\colimgwidth]{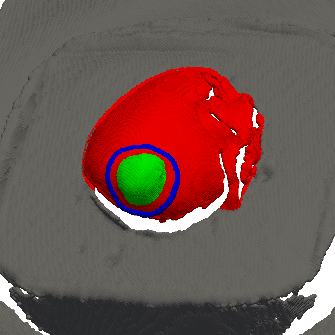}};
        \node[below = 0cm of phantom 4 path.south, anchor = north] (phantom 5 path) {\includegraphics[height=\colimgwidth]{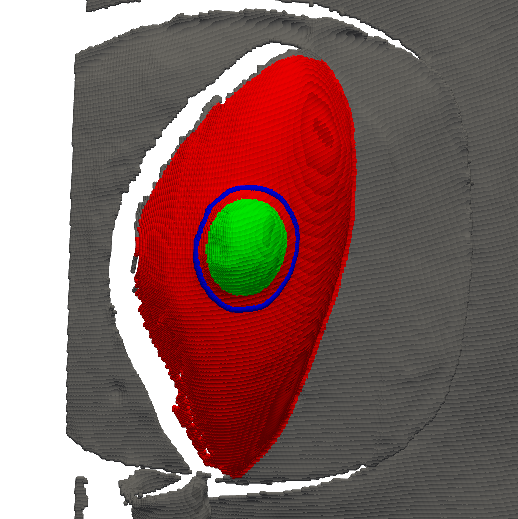}};
        
        \node[right = \imgsep of phantom 1 path.east, anchor=west] (phantom 1 result) {\includegraphics[height=\colimgwidth]{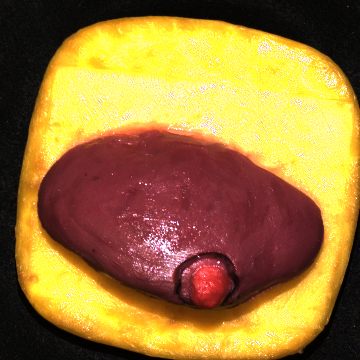}};
        \node[below = 0cm of phantom 1 result.south, anchor = north] (phantom 2 result) {\includegraphics[height=\colimgwidth]{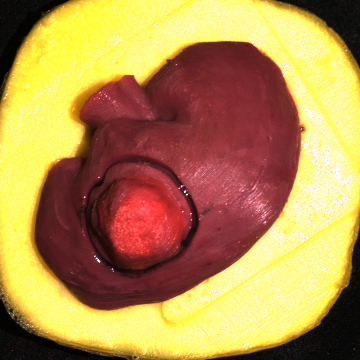}};
        \node[below = 0cm of phantom 2 result.south, anchor = north] (phantom 4 result) {\includegraphics[height=\colimgwidth]{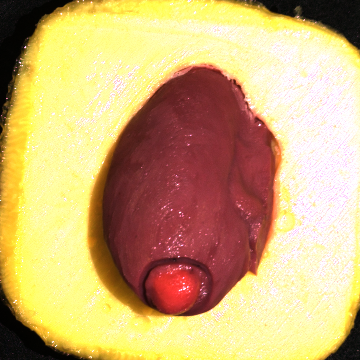}};
        \node[below = 0cm of phantom 4 result.south, anchor = north] (phantom 5 result) {\includegraphics[height=\colimgwidth]{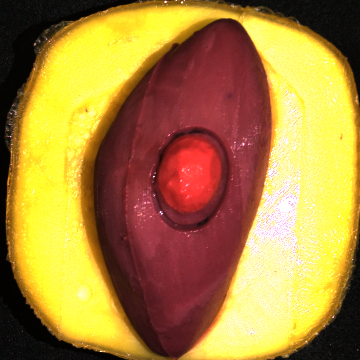}};
        
        \node[right = \imgsep of phantom 1 result.east, anchor=west] (phantom 1 ct) {\includegraphics[height=\colimgwidth]{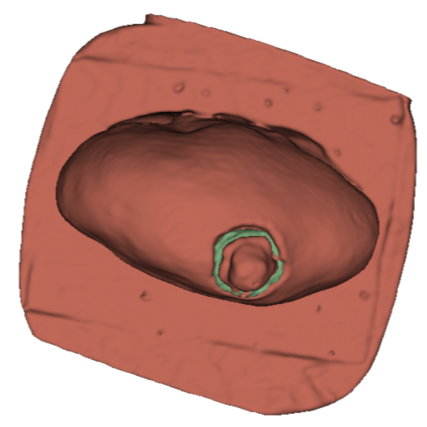}};
        \node[below = 0cm of phantom 1 ct.south, anchor = north] (phantom 2 ct) {\includegraphics[height=\colimgwidth]{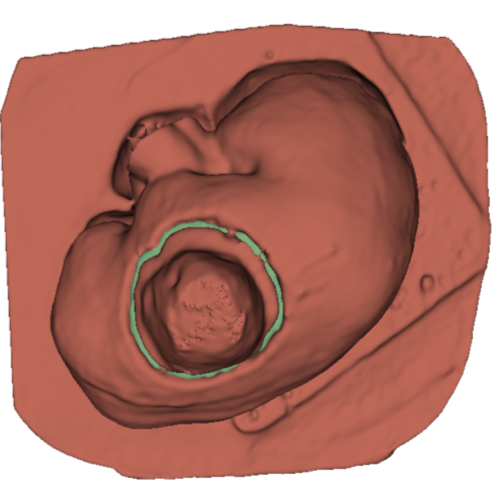}};
        \node[below = 0cm of phantom 2 ct.south, anchor = north] (phantom 4 ct) {\includegraphics[height=\colimgwidth]{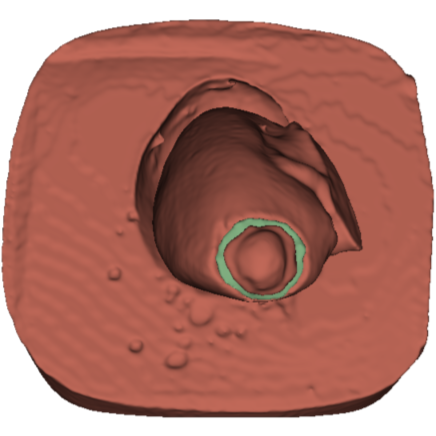}};
        \node[below = 0cm of phantom 4 ct.south, anchor = north] (phantom 5 ct) {\includegraphics[height=\colimgwidth]{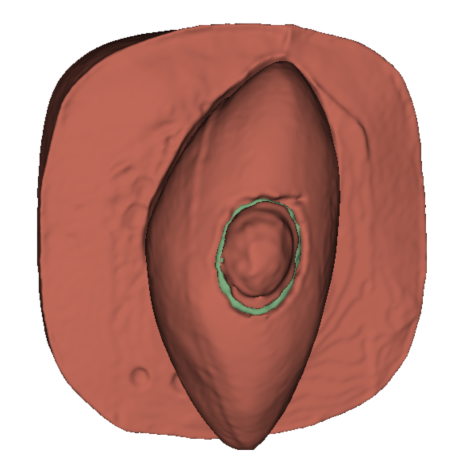}};
% 
        % \node[below = 0.3cm of phantom 1 ct.south, anchor=north, minimum width=\colimgwidth] (phantom 1 size) {\SI{17}{\mm}};
        % \node[below = 0.3cm of phantom 2 ct.south, anchor=north] (phantom 2 size) {\SI{35}{\mm}};
        % %\node[below = 0.3cm of phantom 3 ct.south, anchor=north] (phantom 3 size) {\SI{25}{\mm}};
        % \node[below = 0.3cm of phantom 4 ct.south, anchor=north] (phantom 4 size) {\SI{20}{\mm}};
        % \node[below = 0.3cm of phantom 5 ct.south, anchor=north] (phantom 5 size) {\SI{30}{\mm}};
        % 
        % \node[below = 0.3cm of phantom 1 size.south, anchor=north, minimum width=\colimgwidth] (phantom 1 loc) {below medial};
        % \node[below = 0.3cm of phantom 2 size.south, anchor=north, align=center] (phantom 2 loc) {below anterior\\ upper pole};
        % %\node[below = 0.3cm of phantom 3 size.south, anchor=north] (phantom 3 loc) {left lower pole};
        % \node[below = 0.3cm of phantom 4 size.south, anchor=north] (phantom 4 loc) {below upper pole};
        % \node[below = 0.3cm of phantom 5 size.south, anchor=north] (phantom 5 loc) {below medial};
        % 
% 
        \node[below = 0.2cm of phantom 5.south, anchor = north, font=\bfseries] {Phantom};
        \node[below = 0.2cm of phantom 5 nir.south, anchor = north, font=\bfseries] {NIR Image};
        \node[below = 0.2cm of phantom 5 sam.south, anchor = north, align=center, font=\bfseries] {Segmented\\ NIR Image};
        \node[below = 0.2cm of phantom 5 path.south, anchor = north, align=center, font=\bfseries] {Segmented\\ Point Cloud \&\\ Incision Path};
        \node[below = 0.2cm of phantom 5 result.south, anchor = north, align=center, font=\bfseries] {Executed\\ Incision Path};
        \node[below = 0.2cm of phantom 5 ct.south, anchor = north, align=center, font=\bfseries] {Post Incision\\ CT};
        % \node[left = 0.2cm of phantom 1 size.south, anchor = north, align=center, font=\bfseries] {Tumor Size};
        % \node[left = 0.2cm of phantom 1 loc.south, anchor = north, align=center, font=\bfseries] {Anatomical Site};

        %\draw ($(system.south west) + (-0.5, -0.4)$) -- ($(system.south east) + (0.8, -0.4)$);

        \end{tikzpicture}
        \caption{
        We create four hydrogel phantoms that cover different tumor sizes and anatomical sites.
        The healthy tissue is mixed with a \gls{nir} fluorescent dye to differentiate between healthy and tumorous tissue replicating negative staining \gls{icg}.
        The \gls{nir} image is segmented into background, healthy tissue, and tumor.
        The labels are subsequently used to segment the point cloud observation.
        Based on the segmented point cloud, we plan a robotic incision path that respects a surgical margin around the tumor.
        After incision, we capture a CT scan of the phantom to evaluate incision accuracy.
        }
        \label{fig:models}
    \end{center}
\end{figure*}
\subsection{Autonomous Robotic Margin Delineation and Incision}
The autonomous robotic system consists of a dual-arm robotic setup, a fluorescence-guided perception pipeline to observe the surgical scene, and a planner that determines an accurate incision plan around the tumor, respecting a clinically relevant surgical margin.

\subsubsection{Robotic Experimental Setup}
The dual-arm robotic setup builds upon previous work~\cite{ge2024autonomous} and consists of six main components as shown in  Fig.~\ref{fig1}:
1)~a 6-DOF UR10e manipulator (Universal Robotics, Odense, Denmark) with a custom electrosurgical instrument for tissue dissection, 2) a 6-DOF UR5 manipulator (Universal Robotics, Odense, Denmark) for mounting the camera system, 3) an RGBD camera (Zivid, Oslo, Norway), 4) a 2D \gls{nir} camera (acA2040-90umNIR, Basler AG, Ahrensburg, Germany) with a resolution of $1080\times2048$ together with a 845~$\pm$~27.5~\si{\nano\meter} band-pass filter (Chroma Technology, Bellows Falls, VT), and a \SI{760}{\nano\meter} high-power light-emitting diode (North Coast Technical, Chesterland, OH), and 5) a phantom kidney.
The \gls{nir} camera and light source are selected for their peak excitation and emission wavelengths of \gls{icg} as reported by Ge et al.~\cite{ge2021novel}. 
This setup filters out non-target wavelengths in the image observations, thereby replicating the performance characteristics of \gls{icg}.
A standard monopolar electrode with \SI{25}{\mm} length and \SI{1}{\mm} diameter (Bovie, Clearwater, FL), a grounding pad, and an electrosurgical power generator (ASG-300ESU, DRE Veterinary, Louisville, KY) are used as the electrosurgical instrument.
To filter smoke debris during hydrogel cauterization, we use a a portable smoke evacuator (Smoke Shark, Bovie, Clearwater, FL), and an air purifier (GC Multigas, IQAir, Goldach, Switzerland).

\subsubsection{Perception Pipeline}
Calibration of both camera systems is completed by placing a checkerboard in the scene to serve as a shared world coordinate system.
Following standard hand-eye calibration with the checkerboard, the frames of all robots and cameras are integrated.
For scene reconstruction, we capture a point cloud observation with the RGBD camera and segment the point cloud based on the 2D \gls{nir} image (see second column of Fig.~\ref{fig:models}).
For segmentation, we rely only on the fluorescence and depth point cloud information as renal tumors are often visually indistinguishable from healthy tissue.
We utilize the Segment Anything Model 2 (SAM2)~\cite{ravi2024sam} to segment the \gls{nir} image into classes for background, healthy kidney tissue, and tumor, as shown in the third column of Fig.~\ref{fig:models}.
The frame transformation between the \gls{nir} and depth camera is used to map the observed point cloud into the image space of the \gls{nir} camera.
Each point in the point cloud is labeled as background, healthy kidney tissue, or tumor, based on the segmentation output of the \gls{nir} image from SAM2, as illustrated in the fourth column of Fig.~\ref{fig:models}.

\subsubsection{Incision Path Planner}
In partial nephrectomy, the tumor is resected with a margin of healthy tissue determined by a predefined distance to tumorous tissue.
Based on surgical standards to ensure full tumor removal while minimizing loss of healthy tissue, our incision plan follows a path along the irregular surface of the kidney geometry with a margin of \SI{5}{\mm} to the tumor, following accepted values from clinical practice~\cite{lee2022evaluation}.

The incision planner derives a robotic path using the segmented point cloud from the perception pipeline.
The goal is to identify the incision margin within the segmented point cloud, and plan a path that follows the kidney surface around the outer edge of the incision margin as shown in the fourth column of Fig.~\ref{fig:models}.
To identify the margin, all healthy kidney points that are within \SI{5}{\mm} of any tumor point are labeled as margin points.
The robotic path is derived by first creating a mesh representation of the kidney surface with ball pivoting surface reconstruction, and then identifying the kidney points that connect to the incision margin.

\section{Experiments and Results}
The experiments measure the quality of the proposed realistic kidney phantoms, the accuracy of the proposed perception pipeline, and the overall performance of the autonomous robotic tumor resection system.
With these experiments, we want to answer the following questions.
1) How accurately do the fabricated phantoms replicate the patient anatomy?
2) How good is the \gls{nir} signal from the fluorescent dye inside the phantoms?
3) How accurately can the proposed dual-camera setup produce a segmented scene representation for procedure planning?
4) What is the performance of the executed incision, based on margin accuracy and execution time?

In the following we individually present the experimental setup as well as the results for each component.

\subsection{Kidney Phantom}
The quantitative evaluation of the phantoms addresses 1) the anatomical accuracy of the phantom manufacturing process with respect to the patient data and 2) the quality of the \gls{nir} signal of the fluorescent dye.

    \subsubsection{Anatomical Accuracy}
    Anatomical accuracy is evaluated with the Hausdorff distance between the original patient \gls{ct} images and \gls{ct} images of the manufactured phantoms, captured with a Loop-X system (BrainLab, Germany).
    The Hausdorff distance quantifies the maximum geometric deviation between two point sets, measuring the worst-case difference in spatial correspondence.
    It is defined as
    \begin{equation}
        H(A,B) = \max(h(A,B), h(B,A))
    \end{equation}
    where $h(A,B)$ is the directed Hausdorff distance from set $A$ (\eg, manufactured phantom) to set $B$ (\eg, patient data), given by
    \begin{equation}
        h(A,B) = \max_{a \in A} \min_{b \in B} \| a - b \|
    \end{equation}
    This measures the largest of all distances from a point in $A$ to its nearest neighbor in $B$. 
    This metric provides a robust measure of spatial accuracy between the fabricated phantoms and their intended anatomical geometries.
    The results are presented in the first column of Table~\ref{tab1}.
    The mean Hausdorff distance over all phantoms is \SI{1.5}{\mm}, with minimum and maximum values of \SI{0.82}{\mm} and \SI{2.98}{\mm}.
    
    \subsubsection{Fluorescence Visibility}
    Visibility and contrast of kidney and tumor in \gls{nir} images is measured through the \gls{sbr} metric that evaluates the ratio between target and background intensities:
    \begin{equation}
        \text{SBR} = \frac{\mu_\text{target}}{\mu_\text{background}}
    \end{equation}
    where $\mu_\text{target}$ and $\mu_\text{background}$ represent the mean pixel intensities of the target and background regions, respectively.
    We fabricate three cube-like samples with varying \gls{nir} dye concentrations: 0.38\%, 0.97\%, and 2.04\% (w/w).
    Each sample is cut into five flat equal sections to evaluate internal signal consistency.
    \Gls{sbr} measurements are taken at two time points: immediately after fabrication (Day 1) and after one week of storage (Day 7).
    To reduce fluorescence degradation, all samples are stored at \SI{4}{\degreeCelsius} and shielded from light exposure with aluminum foil wrapping.
    The experimental results are illustrated in Fig.~\ref{fig:sbr}.
    Over the seven-day period, the samples showed an average decline in \gls{sbr} by 29.70\%.
    Additionally, we found a positive linear relation between \gls{nir} dye concentration and \gls{sbr}, showing that the signal intensity of the phantoms can be modulated by adjusting dye concentration.
    
    \begin{tablehere}
            \tbl{Experimental results for anatomical accuracy (column 1) and image segmentation (columns 2 and 3).\label{tab1}}
            {\begin{tabular}{cccc}
            \toprule
            \multirow{2}{*}{Phantom} & Hausdorff & NIR & Point Cloud\\
            & Distance & DICE Score & DICE Score\\ \colrule
            1& \SI{1.26}{\mm} & 0.9715 & 0.7116\\
            2&  \SI{0.82}{\mm} & 0.9676 & 0.9328\\
            3&  \SI{0.93}{\mm} & 0.9695 & 0.6970\\
            4&  \SI{2.98}{\mm} & 0.9532 & 0.8748\\ \colrule
            mean & \SI{1.50}{\mm} & 0.9654 & 0.8041\\
            \botrule
            \end{tabular}}
        \end{tablehere}
    
    \definecolor{OIblack}{RGB}{0, 0, 0}
\definecolor{OIgreen}{RGB}{0, 158, 115}
\definecolor{OIblue}{RGB}{0, 114, 178}
\definecolor{OIlightblue}{RGB}{86, 180, 233}
\definecolor{OIyellow}{RGB}{240, 228, 66}
\definecolor{OIorange}{RGB}{230, 159, 0}
\definecolor{OIred}{RGB}{213, 94, 0}
\definecolor{OIpink}{RGB}{204, 121, 167}

\colorlet{BESO}{OIgreen}
\colorlet{MPD}{OIblue}
\colorlet{DPT}{OIorange}
\colorlet{DPC}{OIpink}
\def\mybarwidth{3mm}
\begin{figurehere}
\centering
\begin{tikzpicture}
\begin{scope}[local bounding box=plot box]
\begin{axis}[
    %axis equal image,
    name=SBRMetrics,
    ybar,
    ymin=0.5,
    ymax=11.9,
    symbolic x coords={Day 1, Day 7},
    xtick=data,
    xlabel={},
    ylabel={SBR},
    ytick distance=3,
    ylabel style={
        at={(-0.4cm, 0.5)},
    trim axis left,
    trim axis right,
    },
    width=0.5\columnwidth,
    height=0.6\columnwidth,
    x tick label style={
        rotate=0,
        anchor=north,
        inner sep=0pt,
        align=center,
        font=\footnotesize,
        text width=1cm,
    },
    y tick label style={
        font=\footnotesize,
    },
    xtick pos=bottom,
    bar width=\mybarwidth,
    ymajorgrids=true,
    yminorgrids=true,
    minor y tick num=1,
    major grid style={thick, black!30!white, dashed},
    minor grid style={ultra thin, black!20!white, dashed},
    major x tick style = transparent,
    minor y tick style = transparent,
    enlarge x limits=0.5,
    legend style={
        overlay,
        at={(0.0,-0.25)},
        anchor=west,
        legend columns=2,
        font=\scriptsize,
        /tikz/every even column/.append style={column sep=0.1cm},
        draw=white,
        },
    legend image code/.code={
        \draw [#1] (0cm,-0.1cm) rectangle (0.15cm,0.2cm); },
    ]
    %\addlegendimage{empty legend}
    %\addlegendentry{\textbf{Concentration:}}

    % 0.38
    \addplot+ [
        fill=MPD,
        draw=MPD!10!black,
        error bars/.cd,
        y dir=both,
        y explicit,
        error bar style={MPD!10!black},
    ] coordinates {
        (Day 1, 6.1142) -= (0.0, 0.3158) += (0.0, 0.2001)
        (Day 7, 3.6548) -= (0.0, 0.2334) += (0.0, 0.1709)
    };
    \addlegendentry{0.38\%};

    % 0.97
    \addplot+ [
        fill=DPC,
        draw=DPC!10!black,
        error bars/.cd,
        y dir=both,
        y explicit,
        error bar style={DPC!10!black},
    ] coordinates {
        (Day 1, 8.3795) -= (0.0, 0.3340) += (0.0, 0.8034)
        (Day 7, 5.826) -= (0.0, 0.7479) += (0.0, 1.0464) 
    };
    \addlegendentry{0.97\%};
    
    % 2.04
    \addplot+ [
        fill=DPT,
        draw=DPT!10!black,
        error bars/.cd,
        y dir=both,
        y explicit,
        error bar style={DPT!10!black},
    ] coordinates {
        (Day 1, 11.1838) -= (0.0, 0.1946) += (0.0, 0.1472)
        (Day 7, 8.5695) -= (0.0, 0.5392) += (0.0, 0.3576)
    };
    \addlegendentry{2.04\%};
\end{axis}

\end{scope}

% Day 1
\node [inner sep = 0, outer sep = 0, anchor=north west] (38 img) at ($(plot box.north east) + (0.3cm, -0.0cm)$)
{\includegraphics[width=0.25\columnwidth]{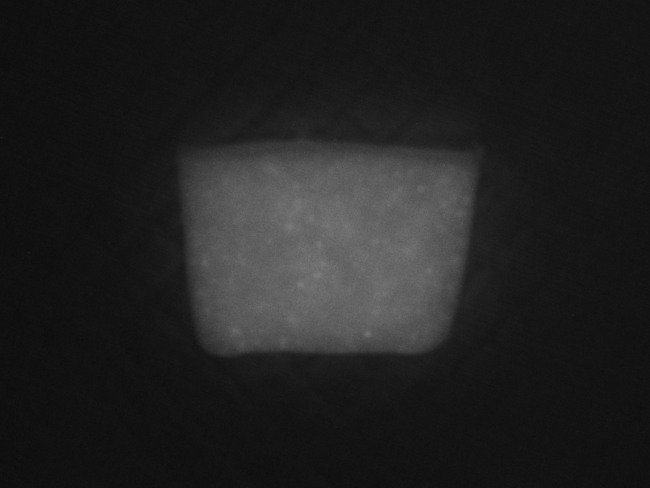}};
\node [inner sep = 0, outer sep = 0, below = 1mm of 38 img.south, anchor=north] (97 img)
{\includegraphics[width=0.25\columnwidth]{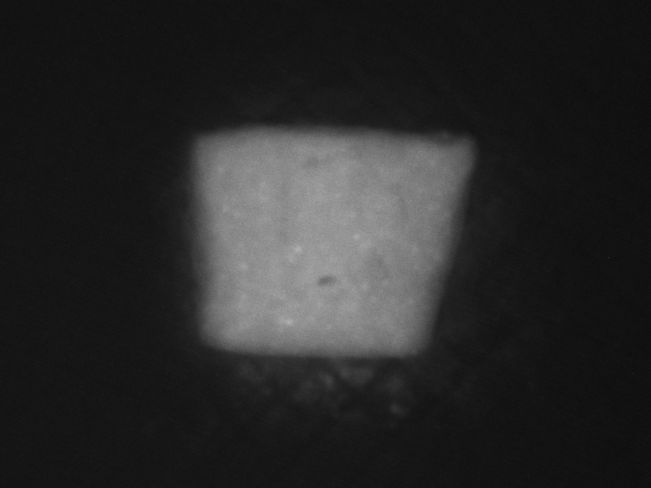}};
\node [inner sep = 0, outer sep = 0, below = 1mm of 97 img.south, anchor=north] (204 img)
{\includegraphics[width=0.25\columnwidth]{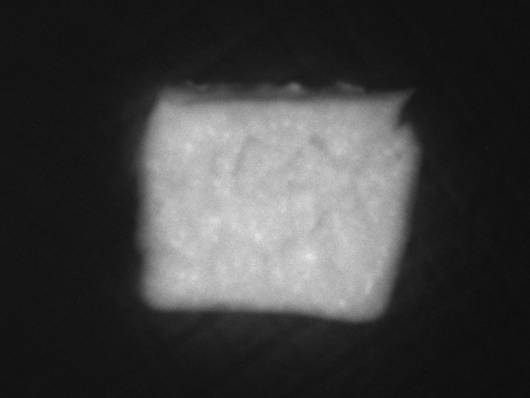}};

% \node [anchor=north west] (204 img) at ($(97 img.north east) + (0, 0.0cm)$) {\includegraphics[width=0.25\columnwidth]{figs/SBR/204_1.png}};
% 
% First day annotations
\node [anchor=north, text=white, outer sep=0cm] at (38 img.north) {Day 1};
\node [anchor=south east, text=white, outer sep=0cm] at (38 img.south east) {0.38\%};
\node [anchor=south east, text=white, outer sep=0cm] at (97 img.south east) {0.97\%};
\node [anchor=south east, text=white, outer sep=0cm] at (204 img.south east) {2.04\%};
% 
% % Second row of images, shifted vertically below the first row
\node [inner sep = 0, outer sep = 0, right = 1mm of 38 img.east, anchor=west] (38 img2)
{\includegraphics[width=0.25\columnwidth]{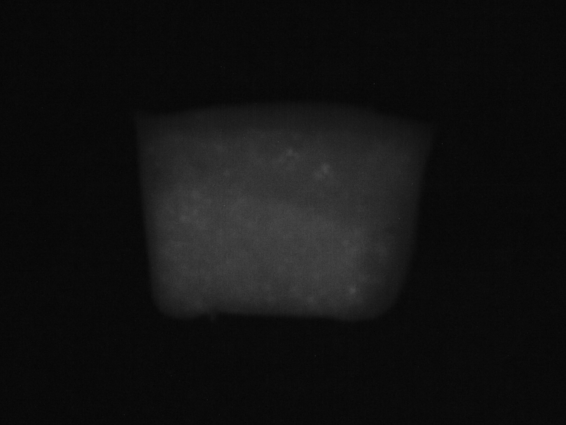}};
\node [inner sep = 0, outer sep = 0, below = 1mm of 38 img2.south, anchor=north ] (97 img2)
{\includegraphics[width=0.25\columnwidth]{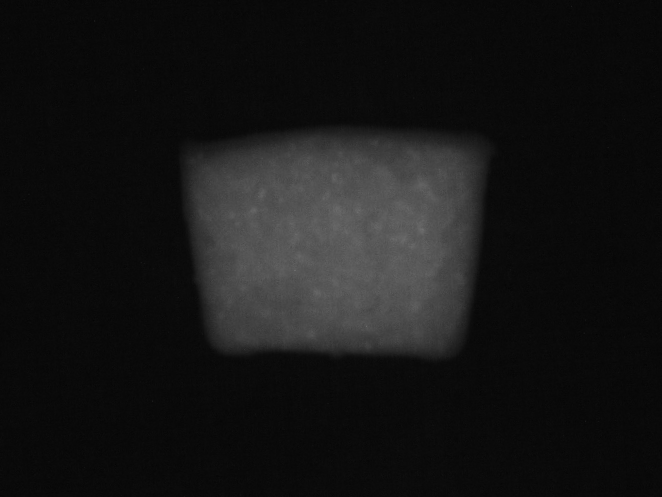}};
\node [inner sep = 0, outer sep = 0, below = 1mm of 97 img2.south, anchor=north] (204 img2)
{\includegraphics[width=0.25\columnwidth]{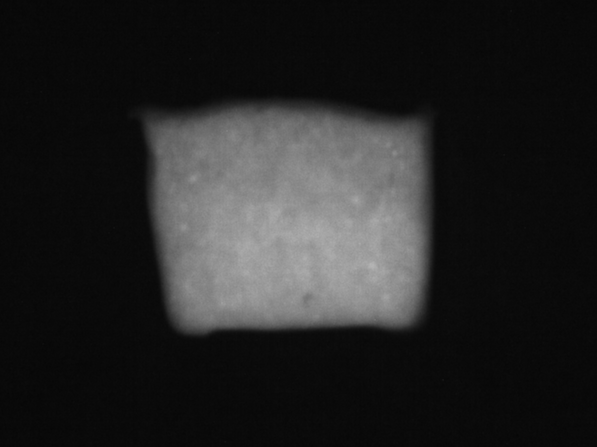}};
% 
% Seventh day annotations
\node [anchor=north, text=white, outer sep=0cm] at (38 img2.north) {Day 7};
\node [anchor=south east, text=white, outer sep=0cm] at (38 img2.south east) {0.38\%};
\node [anchor=south east, text=white, outer sep=0cm] at (97 img2.south east) {0.97\%};
\node [anchor=south east, text=white, outer sep=0cm] at (204 img2.south east) {2.04\%};
% 
% \node [anchor=south, rotate=90] at (38 img.west) {Day 1};
% \node [anchor=south, rotate=90] at (38 img2.west) {Day 7};

\end{tikzpicture}
\caption{
We create samples with different dye concentrations to evaluate the quality of the NIR signal. The correlation between dye concentration and \glsreset{sbr}\gls{sbr} is approximately linear. In cool and shielded storage conditions, the visibility of the dye degrades on average 29.70\% over seven days. 
}
\label{fig:sbr}
\end{figurehere}

\subsection{Autonomous Robotic Margin Delineation and Incision}

    \subsubsection{Experiment Setup}
    Four distinct hydrogel kidney phantoms ($N=4$) are constructed from patient-specific data with varying anatomical features such as tumor size and location.
    The kidney phantoms are glued into sample holders from the same hydrogel material. 
    The sample holders are attached to the table with sutures.
    For the procedure, the camera arm is positioned above the phantom, aligned with surface normal of the tumor, and RGBD and \gls{nir} image samples are taken for scene segmentation.
    The segmented scene is passed to the incision planner for planning and executing an incision path along the kidney surface.

    \subsubsection{Image Segmentation}
    The evaluation of the image segmentation pipeline addresses two individual components: 1) accuracy of the segmentation of the 2D \gls{nir} image and 2) accuracy of the segmented 3D point cloud that potentially compounds errors from inaccurate 2D segmentation and camera-to-camera calibration.
    We employ the \gls{dsc} to quantify segmentation accuracy in both 2D and 3D domains.
    For both cases, \gls{dsc} is calculated as:
    \begin{equation}
        \text{DSC} = \frac{2 |X \cap Y|}{|X| + |Y|}
    \end{equation}
    where $X$ and $Y$ represent the predicted and reference sets respectively, with $|\cdot|$ denoting set cardinality.
    For 2D evaluation, the intersection directly compares pixel labels between automatically and manually annotated masks.
    In 3D point cloud evaluation, we define the intersection using a \SI{0.1}{\mm} proximity threshold, where points are considered matching if their Euclidean distance is smaller than the threshold.
    
    The experimental results are shown in the second and third column of Table~\ref{tab1}.
    Segmentation of the 2D \gls{nir} images achieved consistently high accuracy with \gls{dsc} scores above $0.95$ (mean $0.967\pm0.008$ in range $[0.953, 0.975]$), demonstrating robust algorithm performance.
    Segmentation of the point clouds showed greater variation (mean $0.796\pm0.102$ in range $[0.697, 0.933]$), indicating error propagation from \gls{nir} image segmentation, discretization errors, and camera calibration uncertainties.

    \subsubsection{Autonomous Robotic Electrosurgical Incision}
    The evaluation of the autonomous robotic incision addresses two components: 1) the accuracy of the executed incision with respect to the desired margin and 2) the procedure completion time.
    The executed incisions are shown in the fifth column of Fig.~\ref{fig:models}.
    For each experiment, the power setting of the electrosurgical instrument is set to \SI{40}{\watt}, and tool speed is set to \SI{2}{\mm/\s}.
    To determine the accuracy of the incision, barium sulfate powder is added into the incision for visibility under \gls{ct} as shown in Fig.~\ref{fig:loopx}.
    Each of the four phantoms is \gls{ct} scanned and segmented in 3D Slicer~\cite{3dslicer} to extract the incision in relation to the tumor as illustrated in the final column of Fig.~\ref{fig:models}.
    We calculate the distance between each point in the incision to its closest point on the tumor.
    We report the difference between this distance and the desired distance, specified by the incision margin.
    We further offset the distance by \SI{0.5}{\mm} to account for the diameter of the electrocautery instrument and tissue vaporization during the procedure. 
    The results are presented in Fig.~\ref{fig:ctEval}.
    The system achieves an average of \SI{1.44}{\mm} error with respect to the desired incision margin, and completes the incision within an average of \SI{69}{\sec}.
    
    \begin{figurehere}
        \begin{center}
            \begin{tikzpicture}
                \node[inner sep = 0, outer sep = 0] (loopx) {\includegraphics[height=0.45\columnwidth]{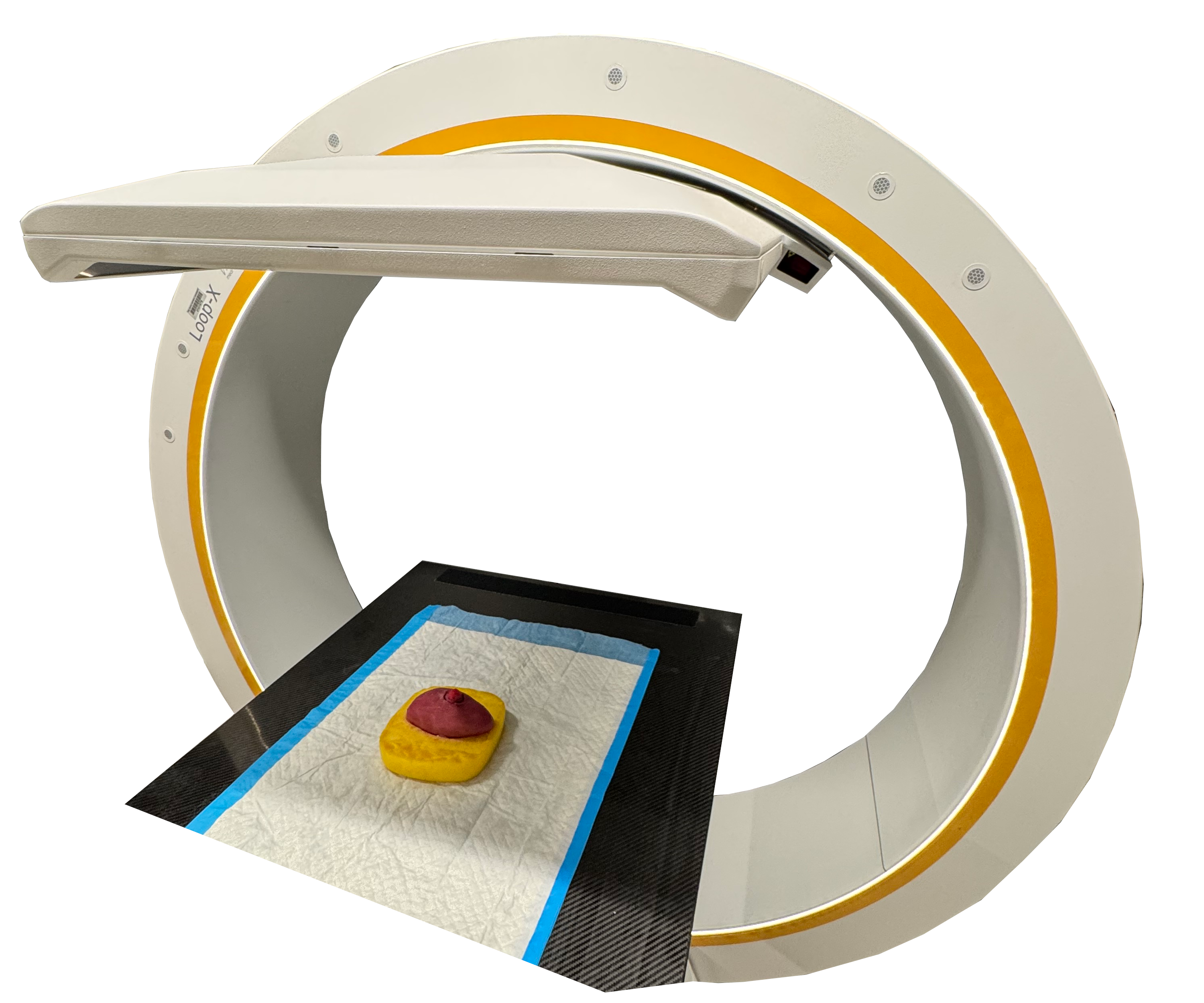}};
                \node[inner sep = 0, outer sep = 0, right = 0cm of loopx.south east, anchor=south west] (barium) {\includegraphics[height=0.4\columnwidth]{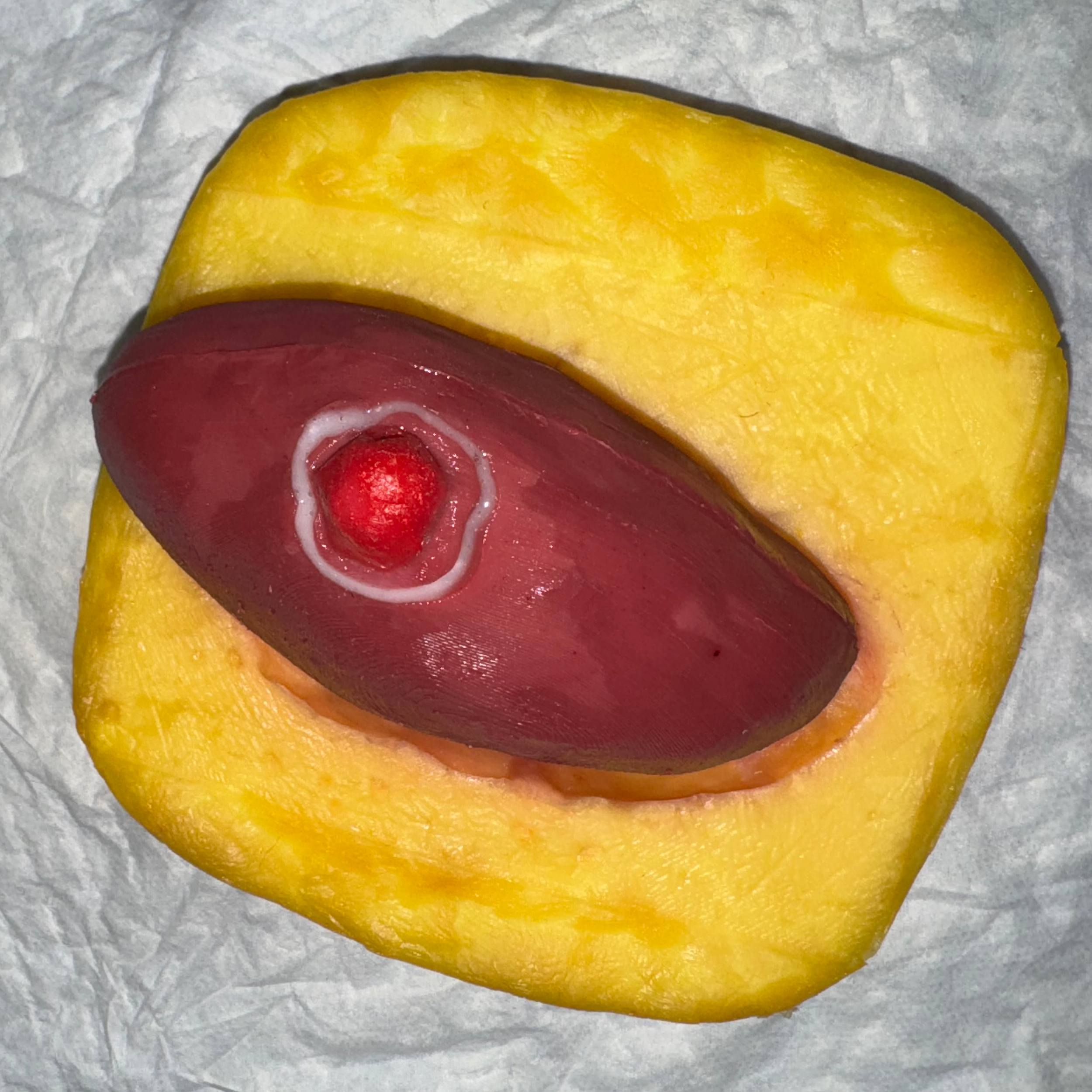}};
                \node[below = 3mm of loopx.south] {a)};
                \node[below = 3mm of barium.south] {b)};
            \end{tikzpicture}
        \end{center}
        \caption{a) After automatic robotic incision, we capture a \gls{ct} scan with a Loop-X system to measure the accuracy of the incision with respect to the desired margin. b) Barium sulfate is added into the incision to visualize the incision in the \gls{ct} scan.}
        \label{fig:loopx}
    \end{figurehere}
    
    \def\colimgwidth{0.25\columnwidth}
\begin{figurehere}
    \centering
    \begin{tikzpicture}
        \node[inner sep=0, outer sep=0.2mm] (ct 6) {\includegraphics[width=\colimgwidth]{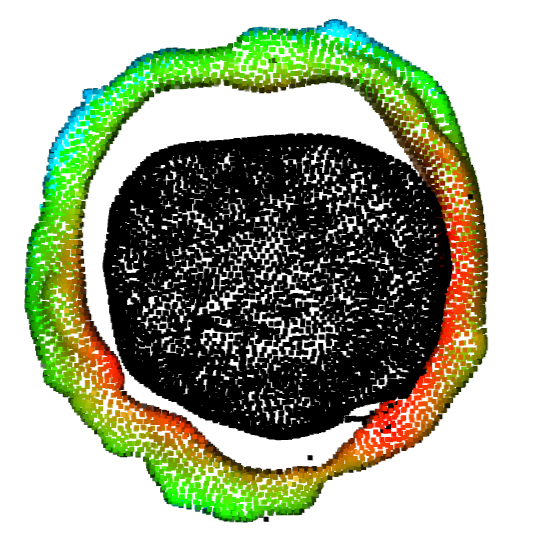}};
        \node [right = 0cm of ct 6.east, anchor = west, inner sep=0, outer sep=0.2mm] (ct 55) {\includegraphics[width=\colimgwidth]{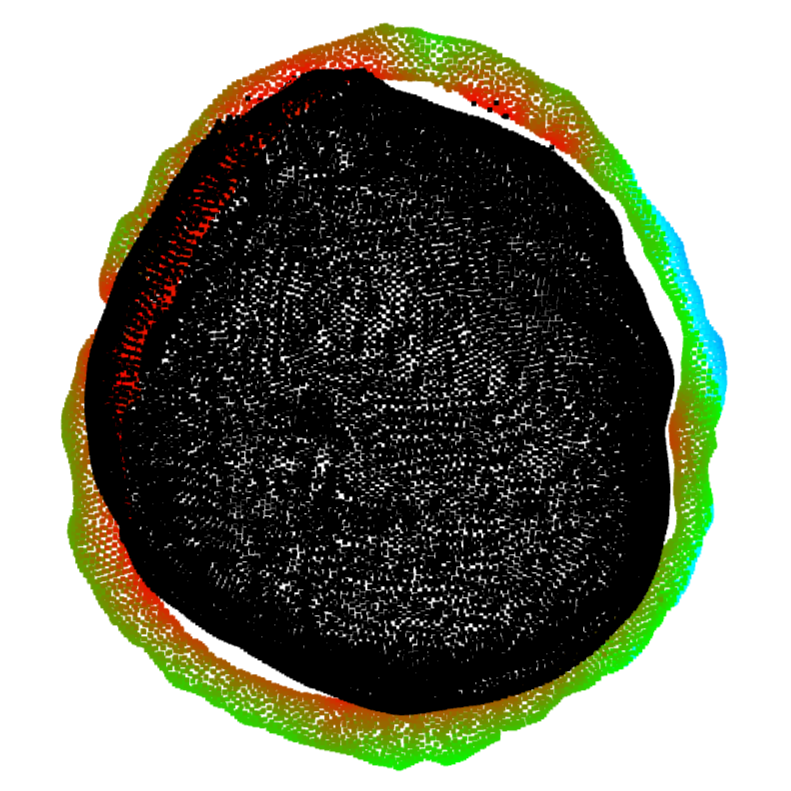}};
        \node [right = 0cm of ct 55.east, anchor = west, inner sep=0, outer sep=0.2mm] (ct 90) {\includegraphics[width=\colimgwidth]{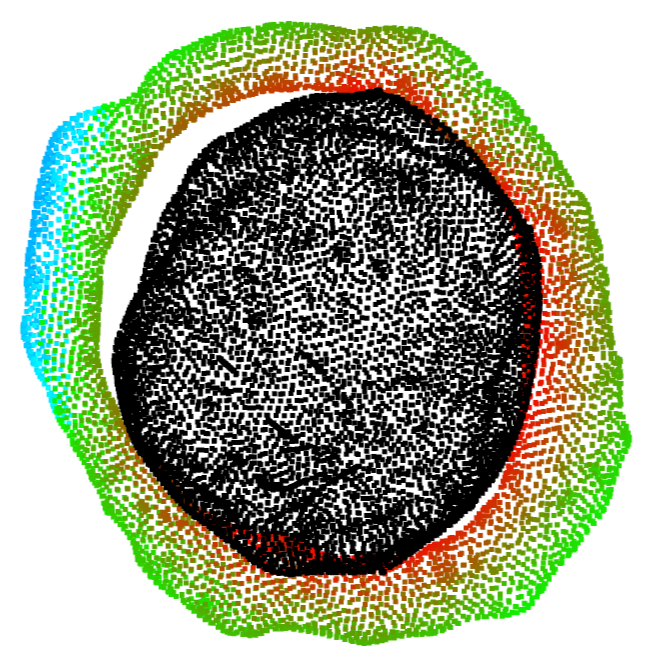}};
        \node [right = 0cm of ct 90.east, anchor = west, inner sep=0, outer sep=0.2mm] (ct 101) {\includegraphics[width=\colimgwidth]{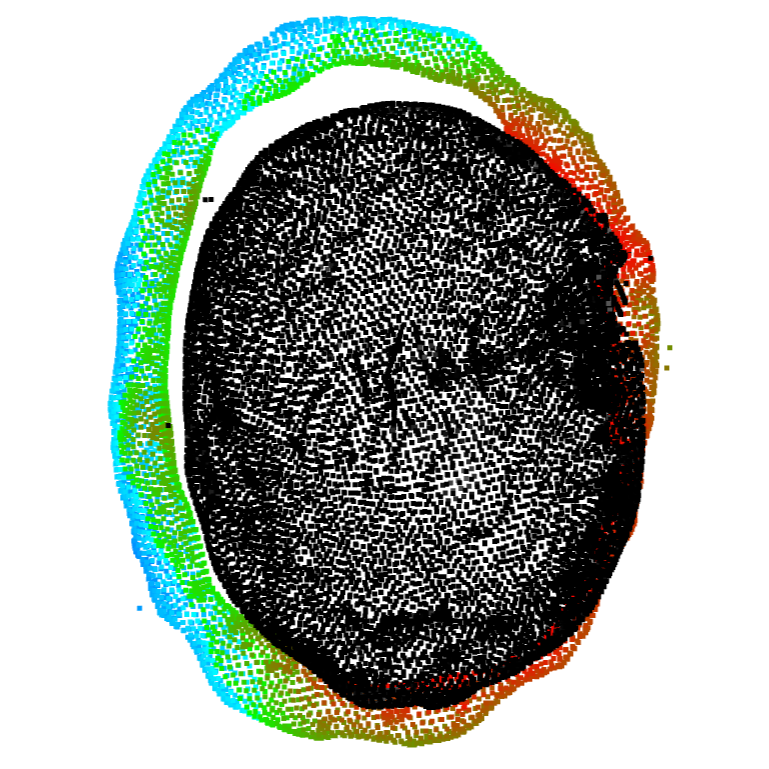}};
        
        \node[below = 0cm of ct 6.south, anchor=north] {\footnotesize $\epsilon: 3.11\pm 2.31$};
        \node[below = 0cm of ct 55.south, anchor=north] {\footnotesize $2.47\pm 1.55$};
        \node[below = 0cm of ct 90.south, anchor=north] {\footnotesize $2.86\pm 1.54$};
        \node[below = 0cm of ct 101.south, anchor=north] {\footnotesize $3.57\pm 2.20$};
    \end{tikzpicture}
    \caption{
        Visualization of tumor (black) and incision segmented from the post-incision CT scans of all four phantoms.
        The incision points are colored relative to their distance to tumor tissue.
        Green values indicate an accurate distance according to the desired resection margin.
        Red points are closer and blue points farther away from tumorous tissue than the desired margin.
        The error values $\epsilon$ are reported as mean $\pm$ standard deviation distance to the desired margin in \si{\mm}.
    }
\label{fig:ctEval}
\end{figurehere}

\section{Discussion}
    \subsection{Translational Limitations}
        The reported hydrogel kidney phantoms allow for testing the fluorescence-guided surgical system.
        However, there are translational limitations between the system and the experimental surgical scene recreation when compared to real-world procedures.
        For example, the kidney phantom does not replicate the full partial nephrectomy scenario, which involves blood and visceral fat occluding the scene, among other anatomy.
        This may introduce boundary ambiguity when delineating the extent of the tumor, a challenge that often occurs in real-world renal cancer surgeries.
        Clinical ICG is dynamic as the human body metabolizes and excretes the dye quickly, giving surgeons only a brief window of visual contrast.
        Further work on our kidney phantoms will attempt to recreate these effects.
        A more complete partial nephrectomy reconstruction would also allow for the evaluation of the SAM2 segmentation method in a more occluded setting and with clinically accurate fluorescence \gls{sbr} levels.
        Moreover, surgical robotics often focus on a \gls{mis} setting, while our setup suggests an open procedure.
        For translation to the \gls{mis} setting, the inclusion of a 3D endoscopic camera would be required, and multiple viewing angles would be necessary for full scene reconstruction of occluded anatomy.

        %Address discrepancy with boundary ambiguity
        %Address discrepancy with moving to MIS
        %Address discrepancy with our scene being too clean
        %Address discrepancy with our scene benefiting from multiple camera angles
        %Address discrepancy with adding additional tool safety (?)
        %Address discrepancy with lower quality images for segmentation

    \subsection{Kidney Phantom}
        The results presented in Fig.~\ref{fig:sbr} show that the \gls{nir} dye is clearly visible in the image.
        Clinically, \gls{sbr} values greater than $2.0$ are considered excellent to distinguish between tumor and healthy tissue~\cite{cancers15030582}.
        In partial nephrectomy, tumors commonly show \gls{sbr} values around $0.5$~\cite{10.3389/fped.2023.1108997}.
        Our reported \gls{sbr} values are considerably higher, which suggests the need for further experiments to find a dye concentration that matches values from clinical reality.
        
        The experimental evaluation of the anatomical accuracy shows good replication of patient anatomy.
        The comparably large deviations for the fourth sample can be accredited to the enlarged Hilar structure in the sample.
        However, these variations in the Hilar structure do not impact the overall accuracy close to the tumorous tissue.

    \subsection{Margin Delineation and Incision}
        The individual components of the proposed perception pipeline show accurate performance in segmenting the tumor from healthy tissue in the kidney phantoms.
        However, relying on observations from a single view limits the overall accuracy for surgical scene reconstruction.
        From a single view, it is challenging to acquire an observation that captures the boundary between tumor and healthy tissue for the complete circumference of the tumor.
        This limitation greatly impacts the ability to accurately plan an incision path.
        In our experiments, we observed that the incisions tended to be closer to the tumor than specified by the desired margin.
        This is due to gaps in the point cloud observation between tumor and healthy tissue caused by the single camera perspective.
        These gaps influence the path planner to select margin points that are too close to the tumor.
        For margin selection, the planner relies on points that are visible from the current camera perspective.
        These observed points are not necessarily points that lie on the surface boundary between tumor and healthy tissue, biasing the planner towards points that are closer to the tumor than desired.
        Additionally, our current robotic system is unable to react to undesired behavior during incision execution from material deformations.
        Force sensing and online trajectory optimization will be included in future work.

    \subsection{Anatomical Limitations}
        Determining a suitable camera perspective becomes more challenging for large exophytic tumors, where the boundary between healthy and tumorous tissue is occluded by the tumor.
        Conversely, our fluorescence-guided perception pipeline is not applicable to very endophytic tumors that are fully embedded in the kidney, as the perception pipeline is unable to observe the negative staining of the tumor from the \gls{nir} image if the tumor is fully within the kidney.

    \subsection{Future Work}
        The complex irregular geometry of our kidney tumors presents a challenge when compared to the simple planar geometries of previous works.
        However, it comes with the additional challenge of adequately visualizing the whole tumor and planning an oriented path along the irregular surface of the kidney that respects the surgical margin.
        Future work will 1) incrementally extend the surgical scene recreation and robotic system to a MIS setting, 2) extend the perception pipeline to reconstruct the surgical scene by merging observations from multiple camera perspectives and 3) improve the autonomous behavior by continuously monitoring incision execution and adapting the plan accordingly.

\section{Conclusion}

%In this study, we showed the feasibility of fluorescence-guided autonomous robotic partial nephrectomy for planning and executing incision paths around exophytic kidney tumors.
% Our contributions include the development of a novel, tissue-mimicking, fluoresence hydrogel kidney phantom derived from patient data, as well as a system for autonomously segmenting tumor and kidney for incision planning and execution with a robotic platform.
%Novel tissue-mimicking fluorescent hydrogel phantoms derived from medically imaged patient data are reported, and we discuss a preception pipeline for fluorescence segmentation of tumor and kidney for incision planning.
%Finally, 

In this work, we present a step towards fluorescence-guided autonomous robotic partial nephrectomy.
Our primary contribution is a robotic system that combines fluorescence-guided imaging for accurate segmentation of renal tumors with autonomous planning and execution of incisions as the first step in a tumor resection procedure.
To facilitate and accelerate development in autonomous robotic tumor resection we present novel, tissue-mimicking, fluorescent hydrogel kidney phantoms.
The phantoms support the use of electrosurgical instruments and are fabricated based on anonymized patient data with various tumor sizes and anatomies.
We report an average incision accuracy of \SI{1.44}{\mm} for the surgical margin and an average completion time of \SI{69}{sec}.
Our future work will focus on autonomously capturing observations from multiple informative camera perspectives to improve scene reconstruction for more complex tumor geometries.

%Our contributions include the development of a novel, tissue-mimicking, fluoresence hydrogel kidney phantom derived from patient data, as well as a system for autonomously segmenting tumor and kidney for incision planning and execution with a robotic platform.

%- In this study, we showed the feasibility of autonomous robotic partial nephrectomy 
%- Our contributions include the development of novel hydrogel-based patient-specific phantoms for kidney sparing nephrectomy
%- a system for autonomously segmenting tumor and kidney to plan and execute an incision path, targeting a key initial component of a partial nephrectomy procedure.
%- We target demonstrating using fluorescence guidance to achieve incisions around a tumor edge on a novel hydrogel phantom
%- We report a novel fluorescence visible hydrogel phantom
%- We use a perception pipeline for scene segmentation 
%- We achieve autonomous robotic partial nephrectomy incision with average margin \SI{1.44}{\mm} and a completion time of \SI{1}{\min} \SI{9}{\sec}.

\nonumsection{Acknowledgments}
\noindent  This material is supported in part by the Advanced Research Projects Agency for Health (ARPA-H) under grant number AY1AX000023, and by the NSF Foundational Research in Robotics (FRR) Faculty Early Career Development Program (CAREER) under grant number 2144348.

\bibliographystyle{ws-jmrr}
\bibliography{jmrr.bib}

\begin{thebibliography}{10}

\bibitem{Siegel2019-iu}
R.~L. Siegel, K.~D. Miller and A.~Jemal, Cancer statistics, 2019, {\em CA: a Cancer Journal for Clinicians} {\bf 69}(1)  (2019)  7--34.

\bibitem{mckinley2016interchangeable}
S.~McKinley {\em et~al.}, An interchangeable surgical instrument system with application to supervised automation of multilateral tumor resection, {\em 2016 IEEE International Conference on Automation Science and Engineering (CASE)\/},  IEEE  (2016), pp. 821--826.

\bibitem{hu2018semi}
D.~Hu, Y.~Gong, E.~J. Seibel, L.~N. Sekhar and B.~Hannaford, Semi-autonomous image-guided brain tumour resection using an integrated robotic system: A bench-top study, {\em The International Journal of Medical Robotics and Computer Assisted Surgery} {\bf 14}(1)  (2018) p. e1872.

\bibitem{ge2024autonomous}
J.~Ge {\em et~al.}, Autonomous system for tumor resection (astr)-dual-arm robotic midline partial glossectomy, {\em IEEE Robotics and Automation Letters} {\bf 9}(2)  (2024)  1166--1173.

\bibitem{marahrens2024ultrasound}
N.~Marahrens, D.~Jones, N.~Murasovs, C.~Biyani and P.~Valdastri, An ultrasound-guided system for autonomous marking of tumor boundaries during robot-assisted surgery, {\em IEEE Transactions on Medical Robotics and Bionics}   (2024).

\bibitem{wang2017review}
K.~Wang, C.-C. Ho, C.~Zhang and B.~Wang, A review on the 3d printing of functional structures for medical phantoms and regenerated tissue and organ applications, {\em Engineering} {\bf 3}(5)  (2017)  653--662.

\bibitem{nieva2024developing}
G.~Nieva-Esteve {\em et~al.}, Developing tuneable viscoelastic silicone gel-based inks for precise 3d printing of clinical phantoms, {\em Materials Advances} {\bf 5}(9)  (2024)  3706--3720.

\bibitem{Amiri2022-rg}
S.~A. Amiri, P.~V. Berckel, M.~Lai, J.~Dankelman and B.~H. Hendriks, Tissue-mimicking phantom materials with tunable optical properties suitable for assessment of diffuse reflectance spectroscopy during electrosurgery, {\em Biomedical Optics Express} {\bf 13}(5)  (2022)  2616--2643.

\bibitem{melnyk2020mechanical}
R.~Melnyk {\em et~al.}, Mechanical and functional validation of a perfused, robot-assisted partial nephrectomy simulation platform using a combination of 3d printing and hydrogel casting, {\em World Journal of Urology} {\bf 38}  (2020)  1631--1641.

\bibitem{abaza2017differential}
R.~Abaza, J.~Rosenthal and J.~Gupta, Differential fluorescence for intraoperative margin assessment with near-infrared fluorescence imaging during robotic partial nephrectomy: Mp52-05, {\em Journal of Urology} {\bf 197}(4)  (2017) p. e705.

\bibitem{henrich2024tracking}
P.~Henrich {\em et~al.}, Tracking tumors under deformation from partial point clouds using occupancy networks, {\em arXiv preprint arXiv:2411.02619}   (2024).

\bibitem{Ghazi2021-ye}
A.~Ghazi {\em et~al.}, Utilizing {3D} printing and hydrogel casting for the development of patient-specific rehearsal platforms for robotic assisted partial nephrectomies, {\em Urology} {\bf 147}  (2021) p. 317.

\bibitem{angell2013optimization}
J.~E. Angell, T.~A. Khemees and R.~Abaza, Optimization of near infrared fluorescence tumor localization during robotic partial nephrectomy, {\em The Journal of urology} {\bf 190}(5)  (2013)  1668--1673.

\bibitem{Krane2012-wn}
L.~S. Krane {\em et~al.}, Is near infrared fluorescence imaging using indocyanine green dye useful in robotic partial nephrectomy: a prospective comparative study of 94 patients, {\em Urology} {\bf 80}(1)  (2012)  110--116.

\bibitem{ge2021novel}
J.~Ge {\em et~al.}, A novel indocyanine green-based fluorescent marker for guiding surgical tumor resection, {\em Journal of Innovative Optical Health Sciences} {\bf 14}(03)  (2021) p. 2150013.

\bibitem{ravi2024sam}
N.~Ravi {\em et~al.}, Sam 2: Segment anything in images and videos, {\em arXiv preprint arXiv:2408.00714}   (2024).

\bibitem{lee2022evaluation}
J.~Lee {\em et~al.}, Evaluation of the surgical margin threshold for avoiding recurrence after partial nephrectomy in patients with renal cell carcinoma, {\em Yonsei Medical Journal} {\bf 63}(2)  (2022) p. 173.

\bibitem{3dslicer}
S.~Pieper, M.~Halle and R.~Kikinis, 3d slicer, {\em IEEE International Symposium on Biomedical Imaging: Nano to Macro\/},   {\bf 1}  (2004), pp. 632--635.

\bibitem{cancers15030582}
M.~F. Gong {\em et~al.}, Intraoperative evaluation of soft tissue sarcoma surgical margins with indocyanine green fluorescence imaging, {\em Cancers} {\bf 15}(3)  (2023) p. 582.

\bibitem{10.3389/fped.2023.1108997}
J.~Feng {\em et~al.}, Clinical application of indocyanine green fluorescence imaging navigation for pediatric renal cancer, {\em Frontiers in Pediatrics} {\bf 11}  (2023) p. 1108997.

\end{thebibliography}

% \noindent\includegraphics[width=1in]{images/author1}
% {\bf Ethan Kilmer} is .\\
% 
% \noindent\includegraphics[width=1in]{images/author2}
% {\bf Joseph Chen} is .
% 
% \noindent\includegraphics[width=1in]{images/author1}
% {\bf Jiawei Ge} is .
% 
% \noindent\includegraphics[width=1in]{images/author1}
% {\bf Richard Cha} is .
% 
% \noindent\includegraphics[width=1in]{images/author1}
% {\bf Kevin Cleary} is .
% 
% \noindent\includegraphics[width=1in]{images/author1}
% {\bf Lauren Shepard} is .
% 
% \noindent\includegraphics[width=1in]{images/author1}
% {\bf Ahmed Ezzat Ghazi} is .
% 
% \noindent\includegraphics[width=1in]{images/author2}
% {\bf Paul Maria Scheikl} is .
% 
% \noindent\includegraphics[width=1in]{images/author1}
% {\bf Axel Krieger} is .
\end{multicols}
\end{document}